\PassOptionsToPackage{table}{xcolor}

\documentclass[10pt,twocolumn,letterpaper]{article}

\usepackage[pagenumbers]{cvpr} 

\usepackage{graphicx}

\makeatletter
\@ifpackagelater{lineno}{2025/12/31}{%
  \AtBeginDocument{%
    \RemoveFromHook{build/column/before}[lineno]%
    \AddToHook{build/column/before}[lineno-fix]{%
      \protected@write\@auxout{}{\string\def\string\@LN@column{\if@firstcolumn2\else1\fi}}}%
  }%
}{}
\makeatother
\usepackage{xcolor}         
\usepackage{caption}
\usepackage{subcaption}
\usepackage{tikz}
\usepackage{placeins}       
\usepackage{sansmath}
\usetikzlibrary{positioning,calc,arrows.meta,shapes.arrows}

\makeatletter
\renewcommand{\paragraph}[1]{\vspace{0.25\baselineskip}\noindent\textbf{#1}\hspace{0.5em}\ignorespaces}
\makeatother

\definecolor{cvprblue}{rgb}{0.21,0.49,0.74}
\usepackage[pagebackref,breaklinks,colorlinks,allcolors=cvprblue]{hyperref}
\usepackage{dsfont}
\def\confName{CVPR}
\def\confYear{2027}

\title{SPiXSeg: Rethinking MLLM-based Image Segmentation as Superpixel Classification}

\author{%
  Piyush Tiwary$^{1}$\quad
  Sohini Dutta$^{2}$\quad
  Prathosh A.P.$^{2}$\\[4pt]
  $^{1}$Dolby Laboratories, Bengaluru\qquad
  $^{2}$Indian Institute of Science, Bengaluru\\[2pt]
  \texttt{piyush.tiwary@dolby.com}\quad
  \texttt{\{sohinidutta, prathosh\}@iisc.ac.in}
}

\begin{document}
\maketitle
\begin{abstract}
Multimodal large language models (MLLMs) are widely used for image segmentation. However, existing methods either use a decoder or use patch-based approach which trades boundary precision against the number of decisions as coarse grid cells mix foreground and background pixels, while finer grids require many more predictions, and autoregressive mask generation adds sequential decoding overhead. We propose \textbf{SPiXSeg}, which shifts the output from per-patch to per-superpixel label assignment, using superpixel tokens formed by aggregating patch-level visual features. An off-the-shelf model partitions each image into task-agnostic superpixels, so the partition is computed once and reused across tasks. A Spix-to-Mask (S2M) module scores each superpixel token in a single MLLM forward pass by directly comparing the MLLM's hidden representations, without autoregressive generation or a separate scoring network, and the selected superpixels are mapped directly to pixels without refinement.
Without any trainable decoder head or post-hoc refinement, SPiXSeg achieves an average cIoU of 80.5 with LLaVA-1.5-7B and 81.9 with InternVL2.5-8B on RefCOCO, RefCOCO+ and RefCOCOg and, when trained only on COCO-Stuff, extends to open-vocabulary semantic segmentation with mIoU of 29.0, 59.5 and 95.4 on ADE20K, PASCAL Context and PASCAL VOC, outperforming several SoTA MLLM baselines. Our results suggest that a fixed, query-agnostic superpixel partition is enough for accurate language-guided segmentation.
\end{abstract}    

\section{Introduction}\label{sec:introduction}
Multimodal Large Language Models (MLLMs) have significantly advanced computer vision by unifying vision and language tasks under a single autoregressive framework~\citep{alayrac2022flamingo,dai2023instructblip,liu2023visual,cocchi2025llava,Yin_2024}, enabling progress across image classification, detection, and visual question answering without task-specific architectural modifications. Extending MLLMs to segmentation requires the mask to be expressed in a form that the model can generate. We refer the unit in which a mask is expressed as the \emph{vocabulary}; essentially it determines the cost of segmentation in terms of the number of decisions, the mask precision, quality, and the modules required before/after the MLLM.

\begin{figure}[t]
\centering
\def\figoneScale{0.75}
\resizebox{\linewidth}{!}{%
  \input{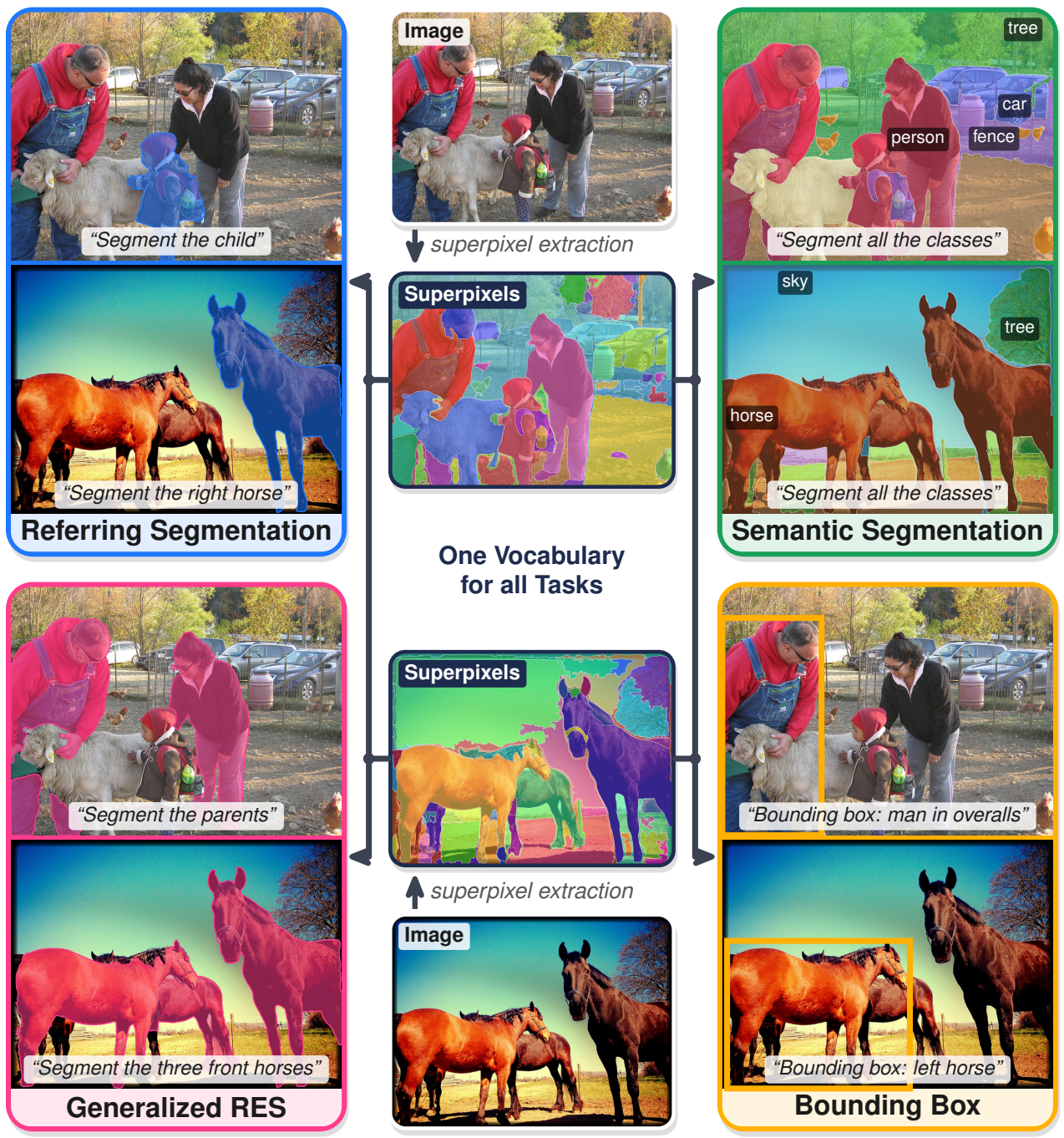}%
}
\caption{We propose superpixels as the primary unit of segmentation. Superpixels are computed once per image, independent of the query, and the same superpixels serve different tasks.}
\label{fig:teaser}
\end{figure}

Existing vocabularies and methods can be broadly grouped into four types. First, methods requiring a decoder after the MLLM, where a special token (e.g., \texttt{<seg>}) is generated, and its hidden state is converted into a mask by a separate decoder~\citep{ren2024pixellm,lai2024lisa,xia2024gsva,rasheed2024glamm,padt}. This adds additional decoder overhead for every query. Second, methods with learned mask tokens, where masks are compressed into discrete tokens by a trained tokenizer~\citep{wang2025himtok,zhou2026samtok}. These methods usually require multi-stage training where a tokenizer is first trained using VQ-VAE~\citep{van2017neural}, then the tokenizer is aligned with the MLLM and then the whole pipeline is trained end-to-end. Third, methods making patch-level predictions, where a label is assigned for every patch of a regular grid~\citep{lantext4seg,stamp,tang2026ufo}. Here, the patch size sets a trade-off between mask quality and sequence length. Large patches produce coarse masks that straddle object boundaries and mix foreground with background, while small patches recover finer detail but require substantially more decisions. Lastly, methods relying on object proposals, where one or more proposals from a detector are selected~\citep{zhang2024groundhog,cao2025RAS}. Proposal-based methods can only output whole detected objects, so they cannot support sub-object segmentation. Moreover, the actual segmentation is done by the detector, and the MLLM only assigns a label, a task that a simple CLIP-style model~\citep{radford2021clip} could handle.

From the above discussion, we emphasize on four properties that are desired from a vocabulary: (i) it should require fewer decisions/predictions per mask/query, (ii) enables precise and accurate masks/boundaries, (iii) should be able to recover mask directly from MLLM, without a decoder, and (iv) manifests a form that can be natively integrated into MLLM and can be reused across queries and various tasks. 
In this work, we observe that superpixels~\cite{achanta2012slic}, which partition an image into non-overlapping regions, satisfy these properties. Our core idea is to use off-the-shelf models to obtain high-quality superpixels for an image and to represent any mask as a union of superpixels. The problem then reduces to integrating superpixels into an MLLM and classifying each superpixel as inside or outside the mask. Since each superpixel groups several pixels into one, the MLLM makes one prediction per superpixel instead of one per pixel. This reduces the number of decisions while retaining accuracy. No additional decoder is needed as the superpixel classifications directly select the relevant superpixels, which are assembled into the final mask. Because superpixels are task-agnostic, they can serve as a general vocabulary for various segmentation tasks, as illustrated in Fig.~\ref{fig:teaser}.
Current literature uses superpixels as input tokens, but are limited to perception tasks like VQA~\citep{chen2025epoc} or fail to natively integrate them for dense tasks like segmentation~\citep{SVP}. In this work, we integrate superpixels natively into MLLM for general-purpose dense segmentation tasks.

\begin{figure*}[t]
\centering
\resizebox{\textwidth}{!}{%
  \input{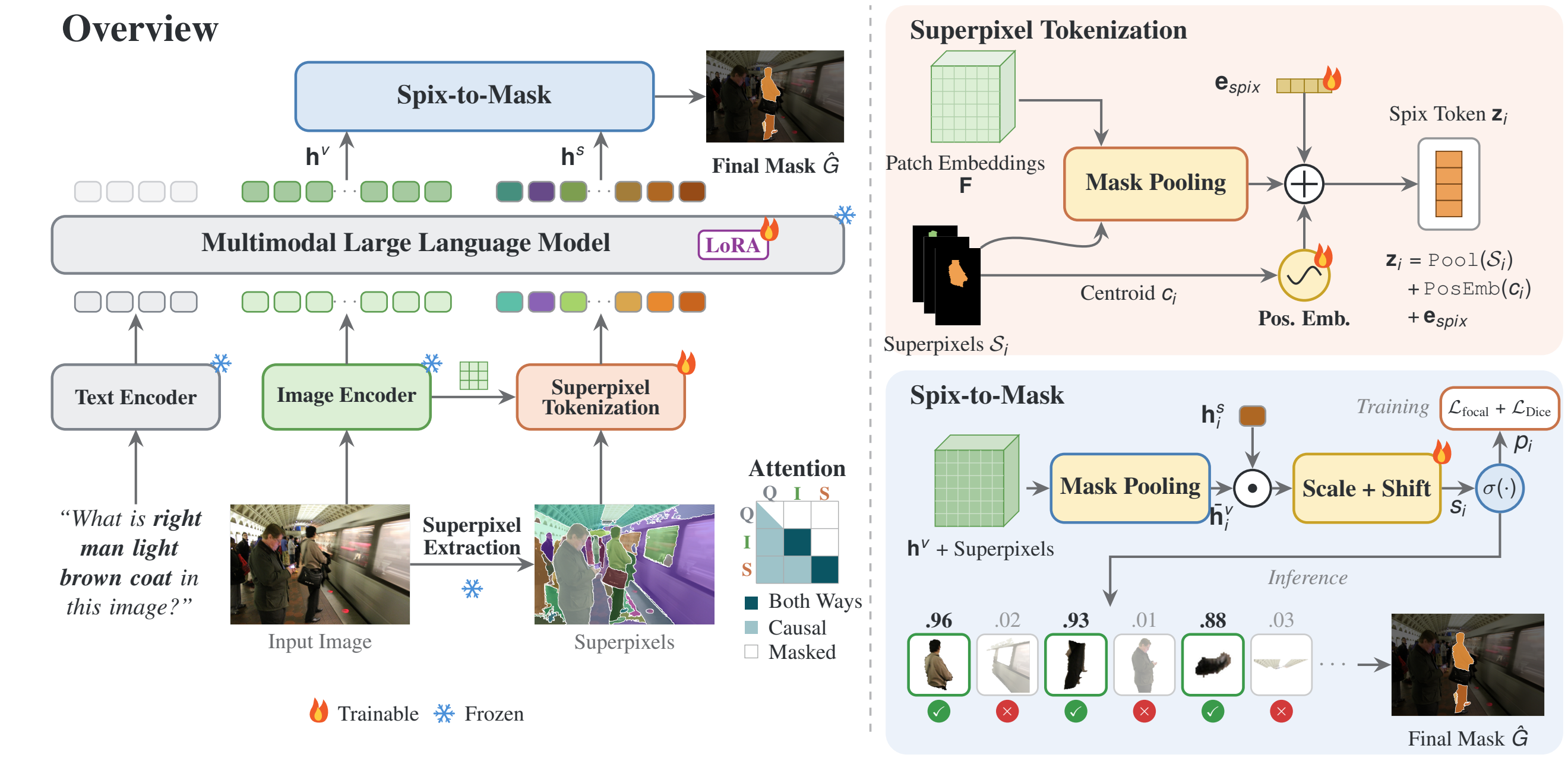}%
}
\caption{Overview of SPiXSeg. The image is partitioned once into superpixels, independent of the query, and each superpixel token is pooled from the MLLM's patch features. S2M scores every superpixel token by comparing hidden states, and the mask is the union of the selected superpixels, with no mask decoder or refinement.}
\label{fig:sparse_overview}
\end{figure*}

Based on this, we propose \textbf{SPiXSeg}. Here, we first extract superpixels from an off-the-shelf model. Next, we compute a token for each superpixel by pooling the patch-level embeddings of the MLLM vision tower over the superpixel region. The superpixel tokens, together with the text and vision tokens, are processed by the transformer backbone of the MLLM in a single forward pass. Then, a Spix-to-Mask (S2M) module classifies each superpixel as inside or outside the mask. S2M does this by comparing the hidden state of the superpixel token with the hidden states of the visual tokens inside that superpixel. This is done without any additional layer on top of the MLLM. Lastly, the mask is obtained as the union of the superpixels classified as inside. An overview of our method is presented in Fig.~\ref{fig:sparse_overview}.

We evaluate SPiXSeg on dense tasks like referring segmentation (RefCOCO/+/g) and open-vocabulary semantic segmentation. With LLaVA-1.5-7B as the backbone, our method achieves a mean cIoU of $80.5$ on referring segmentation, which is $3.5$ to $5.6$ points above the state-of-the-art (SoTA) baselines. Our method also beats baselines with multi-stage training and those that use heavier models like RAS~\cite{cao2025RAS} by significant margin. Further, on semantic segmentation evals, SPiXSeg obtains mIoU of $95.4$ and $59.5$ on PASCAL VOC-20 and PASCAL Context-59 respectively.
In summary, the main contributions are:
\begin{itemize}
  \item We propose \textbf{SPiXSeg} which uses superpixels as the vocabulary for segmentation. Particularly, we recast MLLM segmentation as classification of superpixels. The intuition being that any mask can be represented as union of superpixels, hence segmentation can be boiled down to classification of superpixels as inside or outside the mask. This way, our method is decoder free, precise, and doesn't require any multi-stage training.
  \item Our method first tokenizes each superpixel by aggregating MLLM's own patch-embeddings (Section~\ref{sec:method:tokenization}), the transformer backbone then processes the text, vision and superpixel tokens in a single forward pass. A Spix-to-Mask (S2M) module is introduced, which classifies superpixels by comparing hidden states of the MLLM, with no additional layer added on top (Section~\ref{sec:method:s2m}). 
  \item We obtain SoTA results on RefCOCO/+/g with a mean cIoU of $80.5$ with LLaVA-1.5-7B which is an improvement over existing baselines of about $3.5$ to $5.6$ points. We also show competitive performance on open-vocabulary semantic segmentation and generalized referring segmentation.
\end{itemize}

\section{Related Works}\label{sec:rel_works}

\paragraph{MLLM-based segmentation.}
MLLMs typically consist of a visual encoder and an LLM backbone~\citep{alayrac2022flamingo,dai2023instructblip,liu2023visual,cocchi2025llava,kamath2025gemma3}, these are tied together through a small projector which matches the dimension of visual encoder with LLM backbone. While such architectures have shown great success in perception tasks like VQA~\citep{cocchi2025llava,Zhang2024OMGLLaVABI}, its extension to dense tasks like segmentation remains challenging. Several methods have recently tried to extend MLLMs for segmentation, these can be broadly classified into four types: \emph{(i) Decoder-based} methods, LISA~\citep{lai2024lisa}, GSVA~\citep{xia2024gsva}, PixelLM~\citep{ren2024pixellm}, GLaMM~\citep{rasheed2024glamm}, OMG-LLaVA~\citep{Zhang2024OMGLLaVABI}, NExT-Chat~\citep{zhang2024next} and PaDT~\citep{padt}, use special tokens whose hidden states are turned into a mask by a separate decoder that is run for every query. \emph{(ii) Learned mask tokens}, as in HiMTok~\citep{wang2025himtok} and SAMTok~\citep{zhou2026samtok}, compress a mask into  discrete tokens with a learnt tokenizer and de-tokenizer. This usually leads to multi-stage training for tokenizer, alignment and joint-training. \emph{(iii) Patch-based predictions} where the methods make prediction for each patch of a patchified mask. These predictions can be in form of text as in Text4Seg~\cite{lantext4seg} or in form of special tokens as in STAMP~\citep{stamp,stampplus} and UFO~\citep{tang2026ufo}.
Here the patch size trades mask quality against the number of decisions. A large patch size leads to coarse mask but requires fewer predictions, whereas a smaller patch size leads to accurate mask but requires more predictions. Hence, these methods rely on refiners like SAM or CRF ~\citep{kirillov2023sam,krahenbuhl2011crfs} to convert coarse masks into finegrained outputs. Finally, \emph{(iv) object proposals}, as in Groundhog~\citep{zhang2024groundhog} and RAS~\citep{cao2025RAS}, are produced by a detector and the MLLM selects the relevant mask among them. This limits the output to whole objects, and the overlapping proposals cannot be composed into new masks. Set-of-Mark~\citep{yang2023setofmark} and ChatRex~\citep{jiang2026chatrex} similarly give the MLLM region IDs, but only to refer to regions. In contrast, SPiXSeg expresses the mask in a disjoint, query-agnostic partition of the image, so the MLLM output decodes exactly to a mask without a decoder, tokenizer or refiner.

\paragraph{Superpixels and region tokens.}
Superpixels~\citep{achanta2012slic} group pixels into compact regions that follow image boundaries and partition a given image. Recent off-the-shelf segmenters can produce such partitions at the sub-object level~\citep{kirillov2023sam,carion2025sam3segmentconcepts,chen2025epoc}. Superpixel-like tokens have been used as inputs in several works. E.g., SPiT~\cite{aasan2024spitting} replaces the patch grid of vision transformers with superpixels, EPOC~\citep{chen2025epoc} tokenizes images at the sub-object level for perception tasks such as VQA, and SVP~\citep{SVP} repurposes SAM segments as visual projectors for referring segmentation. These works either target perception or still obtain the mask from a separate decoder. In contrast, our method uses superpixels as the \emph{primary vocabulary} of the MLLM for segmentation. This allows us to re-cast segmentation as a binary classification over superpixels as `\emph{inside}' or `\emph{outside}' the mask.



\section{Methodology}\label{sec:method}
\subsection{Preliminaries and Notation}\label{sec:method:notation}
Let $\mathbf{I} \in \mathbb{R}^{H \times W \times 3}$ denote an input image with spatial resolution $H \times W$, and let $\Omega = \{1, \ldots, H\} \times \{1, \ldots, W\}$ represent its discrete spatial domain.
\paragraph{Task.}Given $\mathbf{I}$ and a text query, the goal is to predict a segmentation mask of $\mathbf{I}$. For referring segmentation, the query is a referring expression, and the output is the mask of the referred region. For semantic segmentation, the query is the class vocabulary $\mathcal{C} = \{c_1, \ldots, c_K\}$ of $K$ classes, and the output is a class label for each pixel.
\paragraph{Superpixel Partitioning.} We define a superpixel partition as a set of $N$ disjoint, contiguous regions $\mathcal{S} = \{\mathcal{S}_1, \ldots, \mathcal{S}_N\}$ that completely cover the image domain:
\begin{align}
  \bigcup_{i=1}^{N} \mathcal{S}_i = \Omega, \qquad \mathcal{S}_i \cap \mathcal{S}_j = \emptyset \quad \forall i \neq j, 
\end{align}
where, $\mathcal{S}_i \subset \Omega$ denotes the set of pixel coordinates comprising the $i$-th superpixel. The spatial extent of each region $\mathcal{S}_i$ is represented by a binary mask $\mathbf{m}_i \in \{0, 1\}^{H \times W}$, where $(\mathbf{m}_i)_{h,w} = 1$ if $(h, w) \in \mathcal{S}_i$, and $0$ otherwise. The superpixels are computed once per image by existing off-the-shelf models~\citep{kirillov2023sam,carion2025sam3segmentconcepts,chen2025epoc}. 
\paragraph{Visual Encoding.} The MLLM's visual encoder processes $\mathbf{I}$ by decomposing it into a regular grid of $P = \lfloor H/p \rfloor \times \lfloor W/p \rfloor$ non-overlapping patches of size $p \times p$. This yields a sequence of patch-level embeddings $\mathbf{F} = \{\mathbf{f}_1, \ldots, \mathbf{f}_P\} \in \mathbb{R}^{P \times d_v}$, where $d_v$ is the visual feature dimension. We denote $\mathcal{P}(\mathcal{S}_i) \subseteq \{1, \ldots, P\}$ as the subset of patch indices whose receptive fields geometrically intersect with superpixel $\mathcal{S}_i$. A projector (usually an MLP) projects the patch-level embeddings to MLLM's input dimension, $d$. 

\subsection{Superpixel Tokenization}\label{sec:method:tokenization}
Once we have the superpixel partitions $\mathcal{S}$ and patch-level embeddings $\mathbf{F}$, we compute the superpixel (spix) tokens by aggregating the patch-level embeddings. Particularly, we first upsample the patch embeddings, $\mathbf{F}$ by a factor of $p$ to match the spatial dimension of $\mathcal{S}$. Then, for $i$-th superpixel $S_i$, we pool the patch embeddings falling inside $\mathcal{S}_i$ by taking a weighted average where the weights are the fraction of the patches lying inside the superpixel:
\begin{align}
  \texttt{Pool}(\mathcal{S}_i) = \sum_{j\in\mathcal{P}(\mathcal{S}_i)} w_{ij}\mathbf{f}_j,~~ w_{ij} = \frac{\texttt{Area}(\mathbf{f}_j \cap \mathcal{S}_i)}{\texttt{Area}(\mathcal{S}_i)}
\label{eq:mask_pool}
\end{align}
Hence, patches lying inside the superpixel contribute more to the embedding compared to ones lying partially inside. Next, to include the positional information to the embedding, we compute the centroid ($c_i$) of $\mathcal{S}_i$ and pass the Fourier positional encodings of $c_i$ through an MLP to get the positional embedding. Lastly, we add a learnable embedding $\mathbf{e}_{spix}$ to differentiate the spix tokens from other tokens. Overall, the spix token $\mathbf{z}_i$ for superpixel $\mathcal{S}_i$ is given by-
\begin{align}
  \mathbf{z}_i = \texttt{Pool}(\mathcal{S}_i) + \texttt{MLP}(\texttt{PosEmb}(c_i)) + \mathbf{e}_{spix}
\label{eq:spix_token}
\end{align}
We note that there is no additional encoding used for spix tokenization, instead we re-use the existing visual patch embeddings to aggregate and re-purpose them for superpixels. This is in contrast with methods like HiMTok~\cite{wang2025himtok}, SAMTok~\cite{zhou2026samtok}, SVP~\cite{SVP} which require separate encoding modules, which even leads to multi-stage pipeline requiring a separate training and aligning of the encoding module. 

\subsection{Spix-to-Mask (S2M)}\label{sec:method:s2m}
Once the spix tokens $\mathbf{Z} = \{\mathbf{z}_1,\cdots,\mathbf{z}_N\}$ are obtained, they are appended to the visual and text tokens, and the entire sequence is processed by the transformer blocks of the MLLM in a single forward pass. The attention mask is designed such that each visual tokens attend each other and the text tokens, whereas each spix tokens attend each other and the visual and text tokens as shown in Fig.~\ref{fig:sparse_overview}.
Let $\mathbf{h}^{v}_j \in \mathbb{R}^{d}$ denote the output hidden state of the $j$-th visual token, and $\mathbf{h}^{s}_i \in \mathbb{R}^{d}$ denote the output hidden state of the $i$-th spix token. 
Since the spix token $\mathbf{z}_i$ attends to both text and visual tokens, $\mathbf{h}^{s}_i$ is conditioned on both. 
Next, we want to classify the spix tokens into the ones `\emph{inside the mask}' and `\emph{outside the mask}'.
Particularly, for a ground-truth mask $G \subset \Omega$, the label of the $i$-th spix token is defined as $y_i = \mathds{1}\!\left[\frac{|\mathcal{S}_i \cap G|}{|\mathcal{S}_i|} > 0.5\right]$, hence, 
 $y_i = 1$ if more than half of $\mathcal{S}_i$ lies inside $G$, and $y_i = 0$ otherwise.
A standard choice for such a classification is a linear layer applied on $\mathbf{h}^{s}_i$. However, such a layer is tied to a fixed set of outputs, and has to decode the content of $\mathcal{S}_i$ from a single vector. Instead, we compare $\mathbf{h}^{s}_i$ against the visual content of $\mathcal{S}_i$, as seen by the MLLM through $\mathbf{h}^{v}_j$. 
To do this, the hidden states of the visual tokens are pooled using the same weights $w_{ij}$ as in Eq.~\ref{eq:mask_pool}- $\bar{\mathbf{h}}^{v}_i = \sum_{j\in\mathcal{P}(\mathcal{S}_i)} w_{ij}\mathbf{h}^{v}_j$.
Hence, $\bar{\mathbf{h}}^{v}_i$ represents what is encoded inside the superpixel location of the image, and $\mathbf{h}^{s}_i$ represents what is inferred about the $i$-th superpixel given the text prompt.
A superpixel is expected to be inside the mask when these two representations/quantities agree. Therefore, we use a simple similarity-based approach for classification.
Particularly, the agreement between $\mathbf{h}^{s}_i$ and $\bar{\mathbf{h}}^{v}_i$ is measured by a scaled dot-product, followed by a sigmoid activation:
\begin{align}
  s_i = \alpha \cdot \frac{\langle \mathbf{h}^{s}_i, \bar{\mathbf{h}}^{v}_i \rangle}{\sqrt{d}} + \beta, \qquad p_i = \sigma(s_i)
\label{eq:s2m_score}
\end{align}
where, $\alpha$ and $\beta$ are learnable scalars, and $p_i \in [0,1]$ is the predicted probability of the spix token being inside the mask. Overall, the classification is performed by comparing two hidden states already produced by the MLLM, and no additional layer is added on top of it. 
This is in contrast with methods requiring a separate decoder where the final mask is obtained by per-pixel prediction or text-encoding based method which requires auto-regressive decoding and suffer through accuracy-latency tradeoff.

\paragraph{Training.} With the labels $y_i$ described above, and the area of each superpixel defined as its number of pixels, $a_i = |\mathcal{S}_i|$, the classification is trained with the sum of area-weighted smooth-Dice loss and Focal loss:
\begin{align}
  \mathcal{L}_{\text{DICE}} &= 1 - \frac{2\sum_{i=1}^{N} a_i\, p_i\, y_i + 1}{\sum_{i=1}^{N} a_i\, p_i + \sum_{i=1}^{N} a_i\, y_i + 1} \label{eq:dice} \\
  \mathcal{L}_{\text{FOCAL}} &= -\frac{1}{N} \sum_{i=1}^{N} \alpha_t \cdot (1 - p_t)^\gamma \cdot \log p_t
\label{eq:focal}
\end{align}
where, $\alpha_t$ is the balancing parameter which equals to $0.25$ if $y_i=1$ and $0.75$ otherwise, $\gamma=2$ is the focusing parameter.
Since the superpixels partition $\Omega$, Eq.~\ref{eq:dice} is identical to the pixel-level Dice between the two maps obtained by assigning $p_i$ and $y_i$ to all pixels in $\mathcal{S}_i$. Hence, the loss is computed over $N$ regions, without upsampling the predictions to the image resolution. 

\begin{figure*}[t]
  \centering
  \begin{minipage}[t]{0.495\textwidth}
    \captionsetup{type=table}
\providecommand{\tbd}{\textcolor{red}{\textbf{?}}}%
\providecommand{\chk}[1]{\textcolor{blue}{#1}}%
\providecommand{\todonum}[1]{\textcolor{orange}{#1}}%
\providecommand{\ours}{SPiXSeg}%
\providecolor{vDec}{HTML}{F0F9F9}%
\providecolor{vTok}{HTML}{F6F3FC}%
\providecolor{vGrid}{HTML}{FDFAE9}%
\providecolor{vProp}{HTML}{FEF5EE}%
\providecolor{vSpix}{HTML}{E9F6E7}%

\centering
\small
\setlength{\tabcolsep}{3.8pt}
\resizebox{\linewidth}{!}{%
\begin{tabular}{lllccccccccc}
\toprule
& & & \multicolumn{3}{c}{\textbf{RefCOCO}} & \multicolumn{3}{c}{\textbf{RefCOCO+}} & \multicolumn{2}{c}{\textbf{RefCOCOg}} & \\
\cmidrule(lr){4-6}\cmidrule(lr){7-9}\cmidrule(lr){10-11}
\textbf{Method} & \textbf{LLM} & \textbf{Mask from} & \textbf{val} & \textbf{testA} & \textbf{testB} & \textbf{val} & \textbf{testA} & \textbf{testB} & \textbf{val} & \textbf{test} & \textbf{Avg}. \\
\midrule
\rowcolor{vDec}\multicolumn{12}{l}{\textit{Decoder after the MLLM}} \\
\rowcolor{vDec}NExT-Chat~\cite{zhang2024next} & Vicuna-7B & SAM & 74.7 & 78.9 & 69.5 & 65.1 & 71.9 & 56.7 & 67.0 & 67.0 & 68.9 \\
\rowcolor{vDec}PixelLM~\cite{ren2024pixellm} & Vicuna-7B & decoder & 73.0 & 76.5 & 68.2 & 66.3 & 71.7 & 58.3 & 69.3 & 70.5 & 69.2 \\
\rowcolor{vDec}LISA~\cite{lai2024lisa} & Vicuna-7B & SAM-H & 74.9 & 79.1 & 72.3 & 65.1 & 70.8 & 58.1 & 67.9 & 70.6 & 69.9 \\
\rowcolor{vDec}LaSagnA~\cite{wei2024lasagna} & Vicuna-7B & SAM-H & 76.8 & 78.7 & 73.8 & 66.4 & 70.6 & 60.1 & 70.6 & 71.9 & 71.1 \\
\rowcolor{vDec}GSVA~\cite{xia2024gsva} & Vicuna-7B & SAM-H & 77.2 & 78.9 & 73.5 & 65.9 & 69.6 & 59.8 & 72.7 & 73.3 & 71.4 \\
\rowcolor{vDec}AnyRef~\cite{he2024anyref} & LLaMA2-7B & SAM-H & 76.9 & 79.9 & 74.2 & 70.3 & 73.5 & 61.8 & 70.0 & 70.7 & 72.2 \\
\rowcolor{vDec}SVP~\cite{SVP} & Vicuna-7B & SAM-H & 80.2 & 82.9 & 77.0 & 71.6 & 76.6 & 65.9 & 73.7 & 74.6 & 75.3 \\
\rowcolor{vDec}GLaMM~\cite{rasheed2024glamm} & Vicuna-7B & SAM-H & 79.5 & 83.2 & 76.9 & 72.6 & 78.7 & 64.6 & 74.2 & 74.9 & 75.6 \\
\rowcolor{vDec}PaDT~\cite{padt} & Qwen2.5-7B & decoder & 78.5 & 79.8 & 77.3 & 75.0 & 77.7 & 71.3 & 73.0 & 73.9 & 75.8 \\

\midrule
\rowcolor{vTok}\multicolumn{12}{l}{\textit{Learned mask tokens}} \\
\rowcolor{vTok}HiMTok~\cite{wang2025himtok} & InternLM2.5-7B & 32 tokens & 81.1 & 81.2 & 79.2 & 77.1 & 78.8 & 71.5 & 75.8 & 76.7 & 77.7 \\
\rowcolor{vTok}SAMTok~\cite{zhou2026samtok} & Qwen2.5-3B & SAM2 & 82.4 & 83.9 & \underline{79.9} & \underline{78.4} & 81.8 & \textbf{74.8} & 79.0 & 79.1 & 79.9 \\
\midrule
\rowcolor{vGrid}\multicolumn{12}{l}{\textit{Grid cells}} \\
\rowcolor{vGrid}Text4Seg~\cite{lantext4seg} & Vicuna-7B$^\ddagger$ & 16$\times$16 grid & 70.5 & 72.3 & 69.3 & 64.4 & 68.7 & 60.6 & 65.1 & 66.5 & 67.2 \\
\rowcolor{vGrid}Text4Seg~\cite{lantext4seg} & Vicuna-7B$^\ddagger$ & grid + CRF & 73.2 & 75.7 & 71.4 & 67.0 & 71.9 & 62.4 & 67.3 & 68.9 & 69.7 \\
\rowcolor{vGrid}Text4Seg~\cite{lantext4seg} & Vicuna-7B$^\ddagger$ & grid + SAM-H & 79.3 & 81.9 & 76.2 & 72.1 & 77.6 & 66.1 & 72.1 & 73.9 & 74.9 \\
\rowcolor{vGrid}STAMP~\cite{stamp} & Vicuna-7B$^\ddagger$ & patches + SAM-H & 80.5 & 83.0 & 77.3 & 74.8 & 79.2 & 68.6 & 75.7 & 76.8 & 77.0 \\
\rowcolor{vGrid}UFO~\cite{tang2026ufo} & Vicuna-7B$^\ddagger$ & per-pixel & 77.2 & 80.1 & 76.4 & 71.8 & 77.9 & 70.2 & 74.1 & 73.5 & 75.2 \\
\rowcolor{vGrid}UFO (RES ft)~\cite{tang2026ufo} & Vicuna-7B$^\ddagger$ & per-pixel & 77.9 & 81.1 & 77.0 & 72.5 & 78.5 & 71.4 & 75.6 & 74.1 & 76.0 \\
\rowcolor{vGrid}UFO (RES ft)~\cite{tang2026ufo} & InternLM2.5-7B & per-pixel & 81.0 & 82.6 & 78.6 & 77.1 & 80.4 & 72.6 & 76.7 & 77.3 & 78.3 \\
\midrule
\rowcolor{vProp}\multicolumn{12}{l}{\textit{Object proposals}} \\
\rowcolor{vProp}RAS~\cite{cao2025RAS} & Vicuna-7B$^\ddagger$ & Co-DETR & -- & -- & -- & -- & -- & -- & -- & -- & 73.7 \\
\rowcolor{vProp}Groundhog~\cite{zhang2024groundhog} & LLaMA2-7B & Mask2Former+ & 78.5 & 79.9 & 75.7 & 70.5 & 75.0 & 64.9 & 74.1 & 74.6 & 74.2 \\
\rowcolor{vProp}RAS~\cite{cao2025RAS} & Vicuna-13B$^\ddagger$ & Co-DETR & 79.4 & 82.6 & 75.9 & 72.2 & 77.3 & 64.7 & 73.2 & 74.5 & 75.0 \\
\rowcolor{vProp}RAS (RES ft)~\cite{cao2025RAS} & Vicuna-13B$^\ddagger$ & Co-DETR & 81.0 & 83.5 & 79.0 & 75.1 & 80.0 & 70.3 & 76.0 & 77.5 & 77.8 \\
\midrule
\rowcolor{vSpix}\multicolumn{12}{l}{\textit{Superpixels}} \\
\rowcolor{vSpix}\textbf{\ours{} (ours)} & Vicuna-7B$^\ddagger$ & superpixels & \underline{84.1} & \underline{86.9} & 79.7 & 78.2 & \underline{83.7} & 71.0 & \underline{79.5} & \underline{80.8} & \underline{80.5} \\
\rowcolor{vSpix}\textbf{\ours{} (ours)} & InternLM2.5-7B & superpixels & \textbf{84.5} & \textbf{87.1} & \textbf{81.3} & \textbf{80.5} & \textbf{84.9} & \underline{74.2} & \textbf{80.6} & \textbf{81.9} & \textbf{81.9} \\
\bottomrule
\end{tabular}%
}

    \captionof{table}{Referring expression segmentation (cIoU), grouped by the unit in which the mask is expressed. \ours{} uses no mask decoder or refiner. Bold / underline: best / second best. $^\ddagger$LLaVA-1.5. ft: fine-tuned per dataset. RAS-7B reports only the average.}
    \label{tab:res}
  \end{minipage}\hfill
  \begin{minipage}[t]{0.495\textwidth}
    \captionsetup{type=table}
\providecommand{\tbd}{\textcolor{red}{\textbf{?}}}%
\providecommand{\chk}[1]{\textcolor{blue}{#1}}%
\providecommand{\todonum}[1]{\textcolor{orange}{#1}}%
\providecommand{\ours}{SPiXSeg}%
\providecolor{vDec}{HTML}{F0F9F9}%
\providecolor{vTok}{HTML}{F6F3FC}%
\providecolor{vGrid}{HTML}{FDFAE9}%
\providecolor{vProp}{HTML}{FEF5EE}%
\providecolor{vSpix}{HTML}{E9F6E7}%

\centering
\small
\setlength{\tabcolsep}{3.8pt}
\resizebox{\linewidth}{!}{%
\begin{tabular}{lllccccccccc}
\toprule
& & & \multicolumn{3}{c}{\textbf{RefCOCO}} & \multicolumn{3}{c}{\textbf{RefCOCO+}} & \multicolumn{2}{c}{\textbf{RefCOCOg}} & \\
\cmidrule(lr){4-6}\cmidrule(lr){7-9}\cmidrule(lr){10-11}
\textbf{Method} & \textbf{LLM} & \textbf{Box from} & \textbf{val} & \textbf{testA} & \textbf{testB} & \textbf{val} & \textbf{testA} & \textbf{testB} & \textbf{val} & \textbf{test} & \textbf{Avg.} \\
\midrule
\multicolumn{12}{l}{\textit{Specialist models}} \\
MDETR~\cite{mdetr} & -- & box head & 86.8 & 89.6 & 81.4 & 79.5 & 84.1 & 70.6 & 81.6 & 80.9 & 81.8 \\
Grounding DINO-L~\cite{groundingdino} & -- & box head & 90.6 & 93.2 & 88.2 & 82.8 & 89.0 & 75.9 & 86.1 & 87.0 & 86.6 \\
UNINEXT-L~\cite{uninext} & -- & box head & 91.4 & 93.7 & 88.9 & 83.1 & 87.9 & 76.2 & 86.9 & 87.5 & 87.0 \\
\midrule
\multicolumn{12}{l}{\textit{MLLMs, box output}} \\
Shikra~\cite{shikra} & Vicuna-7B & text & 87.0 & 90.6 & 80.2 & 81.6 & 87.4 & 72.1 & 82.3 & 82.2 & 82.9 \\
Ferret~\cite{you2024ferret} & Vicuna-7B & text & 87.5 & 91.4 & 82.5 & 80.8 & 87.4 & 73.1 & 83.9 & 84.8 & 83.9 \\
Qwen-VL-Chat~\cite{qwenvl} & Qwen-7B & text & 88.6 & 92.3 & 84.5 & 82.8 & 88.6 & 76.8 & 86.0 & 86.3 & 85.7 \\
UFO~\cite{tang2026ufo} & Vicuna-7B$^\ddagger$ & text & 90.2 & 93.5 & 87.3 & 84.4 & 90.3 & 78.7 & 86.4 & 86.8 & 87.2 \\
UFO (ft)~\cite{tang2026ufo} & Vicuna-7B$^\ddagger$ & text & 91.1 & 93.7 & 88.6 & 85.5 & 90.5 & 79.9 & 87.3 & 87.2 & 88.0 \\
\midrule
\rowcolor{vDec}\multicolumn{12}{l}{\textit{Decoder after the MLLM}} \\
\rowcolor{vDec}LISA~\cite{lai2024lisa} & Vicuna-7B & SAM-H  & 85.4 & 88.8 & 82.6 & 74.2 & 79.5 & 68.4 & 79.3 & 80.4 & 79.8 \\
\rowcolor{vDec}GSVA~\cite{xia2024gsva} & Vicuna-7B & SAM-H  & 86.3 & 89.2 & 83.8 & 72.8 & 78.8 & 68.0 & 81.6 & 81.8 & 80.3 \\
\rowcolor{vDec}PixelLM~\cite{ren2024pixellm} & Vicuna-7B & Decoder & 89.8 & 92.2 & 86.4 & 83.2 & 87.0 & 78.9 & 84.6 & 86.0 & 86.0 \\
\rowcolor{vDec}PaDT~\cite{padt} & Qwen2.5-7B & Decoder & 93.1 & 97.2 & \underline{90.4} & 88.8 & 92.8 & 83.2 & 88.2 & 88.8 & 90.1 \\
\midrule
\rowcolor{vTok}\multicolumn{12}{l}{\textit{Learned mask tokens}} \\
\rowcolor{vTok}HiMTok~\cite{wang2025himtok} & InternLM2.5-7B & text after mask tokens & 92.9 & 94.7 & 89.3 & 87.6 & 91.5 & 81.5 & 88.5 & 89.0 & 89.4 \\
\rowcolor{vTok}SAMTok~\cite{zhou2026samtok} & Qwen2.5-3B & SAM2  & 92.7 & 94.6 & 89.7 & 88.2 & 92.2 & \underline{84.4} & \underline{89.9} & 89.6 & 90.2 \\
\midrule
\rowcolor{vGrid}\multicolumn{12}{l}{\textit{Grid cells}} \\
\rowcolor{vGrid}Text4Seg~\cite{lantext4seg} & Vicuna-7B$^\ddagger$ & grid mask & 89.2 & 92.0 & 86.4 & 83.4 & 88.6 & 78.0 & 81.7 & 82.4 & 85.2 \\
\rowcolor{vGrid}Text4Seg~\cite{lantext4seg} & Vicuna-7B$^\ddagger$ & grid + SAM-H & 90.8 & 93.7 & 87.6 & 84.7 & 90.2 & 79.0 & 84.8 & 85.0 & 87.0 \\
\midrule
\rowcolor{vProp}\multicolumn{12}{l}{\textit{Object proposals}} \\
\rowcolor{vProp}Groma~\cite{ma2024groma} & Vicuna-7B & Proposals & 89.5 & 92.1 & 86.3 & 83.9 & 88.9 & 78.1 & 86.4 & 87.0 & 86.5 \\
\rowcolor{vProp}ChatRex~\cite{jiang2026chatrex} & Qwen2.5-7B & Proposals & 91.0 & 94.1 & 87.0 & \underline{89.8} & 91.9 & 79.3 & 89.8 & 90.0 & 89.1 \\
\midrule
\rowcolor{vSpix}\multicolumn{12}{l}{\textit{Superpixels}} \\
\rowcolor{vSpix}\textbf{\ours{} (ours)} & Vicuna-7B$^\ddagger$ & Superpixels & \underline{94.6} & \textbf{98.2} & 89.1 & 89.7 & \underline{94.2} & 82.3 & 89.0 & \underline{90.1} & \underline{90.9} \\
\rowcolor{vSpix}\textbf{\ours{} (ours)} & InternLM2.5-7B & Superpixels & \textbf{94.8} & \underline{98.1} & \textbf{90.7} & \textbf{91.0} & \textbf{95.7} & \textbf{85.2} & \textbf{90.9} & \textbf{91.4} & \textbf{92.2} \\
\bottomrule
\end{tabular}%
}

    \captionof{table}{Referring expression comprehension (Acc@0.5). Segmentation models are scored with the box of their predicted mask (PaDT: decoder box; HiMTok: box written as text; Groma, ChatRex: selected proposal); \ours{} uses the RES model. Bold / underline: best / second best. $^\ddagger$LLaVA-1.5. ft: fine-tuned.}
    \label{tab:rec}
  \end{minipage}
\end{figure*}

\paragraph{Inference.} During inference, the $i$-th superpixel is classified as inside the mask if $p_i > 0.5$, for the corresponding token, and the predicted mask is given by the union of such superpixels, $\hat{G} = \bigcup_{i:\, p_i > 0.5} \mathcal{S}_i$.
As the superpixels are disjoint and cover $\Omega$, any classification of the spix tokens is converted into a mask exactly without any refinement.

\paragraph{Semantic Segmentation.} For semantic segmentation, the classification is extended from two outcomes (inside and outside) to the $K$ classes of $\mathcal{C}$. Let $\mathbf{h}_c \in \mathbb{R}^d$ denote the hidden state of the name of class $c$ in the text tokens. If there are multiple tokens corresponding to the class name, we take the average. The score of class $c$ for $\mathcal{S}_i$ is computed by comparing $\mathbf{h}_c$ to both $\mathbf{h}^s_i$ and $\bar{\mathbf{h}}^v_i$:
\begin{equation}
  s_{i,c} = \tau \cdot \big(\cos(\mathbf{h}_c, \mathbf{h}^s_i) + \cos(\mathbf{h}_c, \bar{\mathbf{h}}^v_i)\big)
\end{equation}
where $\tau$ is a learnable scalar. As a superpixel may contain several classes, its label $y_{i,c} = |\mathcal{S}_i \cap G_c| / |\mathcal{S}_i|$ is the fraction of $\mathcal{S}_i$ belonging to class $c$, with $G_c \subset \Omega$ the ground-truth region of $c$. With $p_{i,c} = \operatorname{softmax}_{c \in \mathcal{C}}(s_{i,c})$, the loss is a soft cross-entropy plus the Dice loss of Eq.~\ref{eq:dice} applied per class:
\begin{equation}
\begin{split}
  \mathcal{L}_{\text{SEM}} = -\frac{1}{N}\sum_{i=1}^{N}\sum_{c \in \mathcal{C}} y_{i,c}\log p_{i,c} 
  + \frac{1}{K}\sum_{c \in \mathcal{C}} \mathcal{L}^{c}_{\text{DICE}},
\end{split}
\end{equation}
where $\mathcal{L}^{c}_{\text{DICE}}$ is Eq.~\ref{eq:dice} with $p_{i,c}$ and $y_{i,c}$ in place of $p_i$ and $y_i$. Then, the predicted label of $\mathcal{S}_i$ is $\hat{y}_i = \arg\max_{c \in \mathcal{C}} s_{i,c}$, which is assigned to all pixels in $\mathcal{S}_i$. Hence, referring and semantic segmentation share the same mechanism, and differ only in what the hidden states are compared against.

\begin{figure*}[t]
  \centering
  \begin{subfigure}[t]{0.495\textwidth}
    \centering
    \vspace{0pt}
    \resizebox{\linewidth}{!}{\input{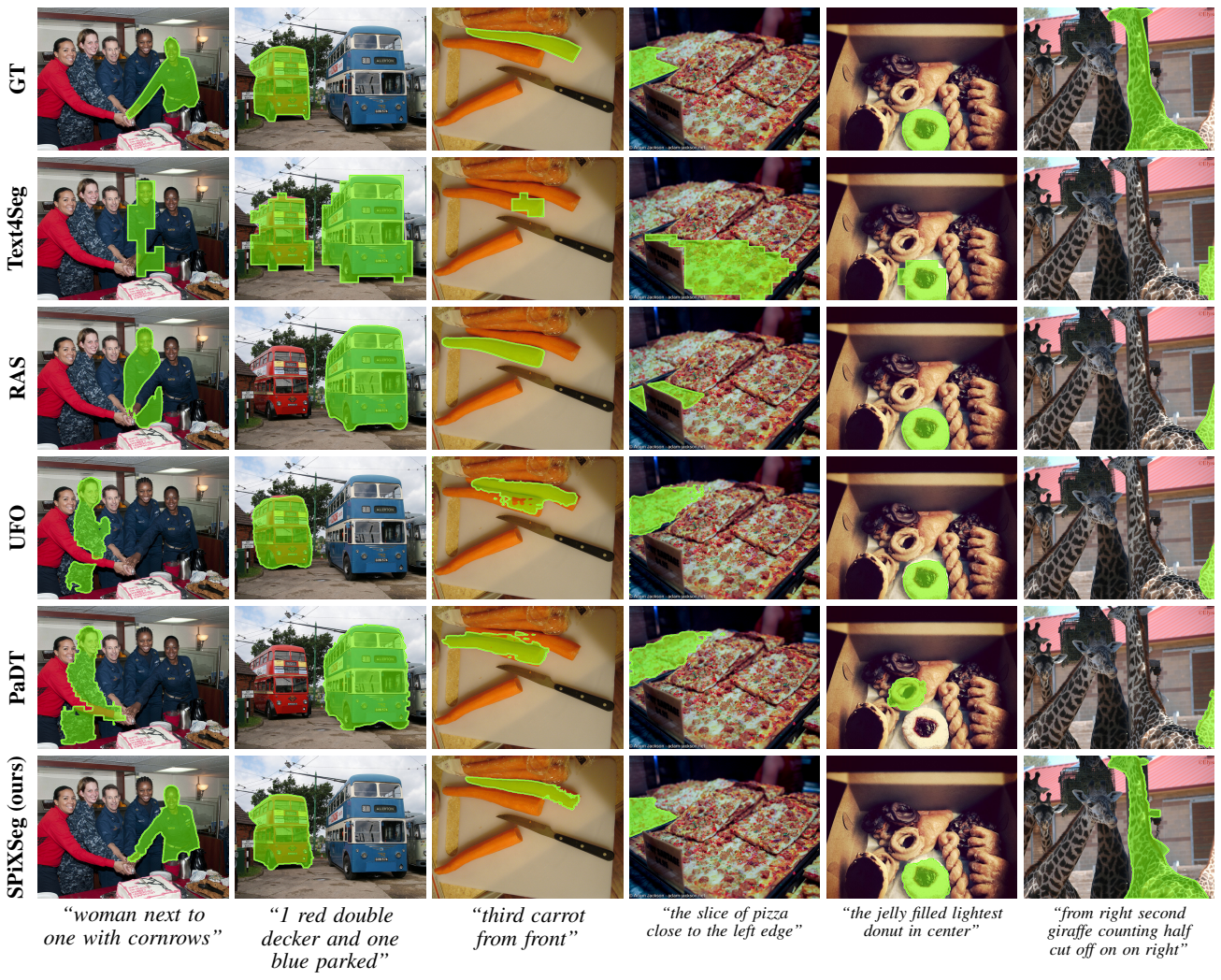}}
    \caption{\textbf{RES.} Ground-truth masks (GT) and predictions from representative baselines and SPiXSeg on RefCOCO.}
    \label{fig:res_compare}
  \end{subfigure}\hfill
  \begin{subfigure}[t]{0.495\textwidth}
    \centering
    \vspace{0pt}
    \resizebox{\linewidth}{!}{\input{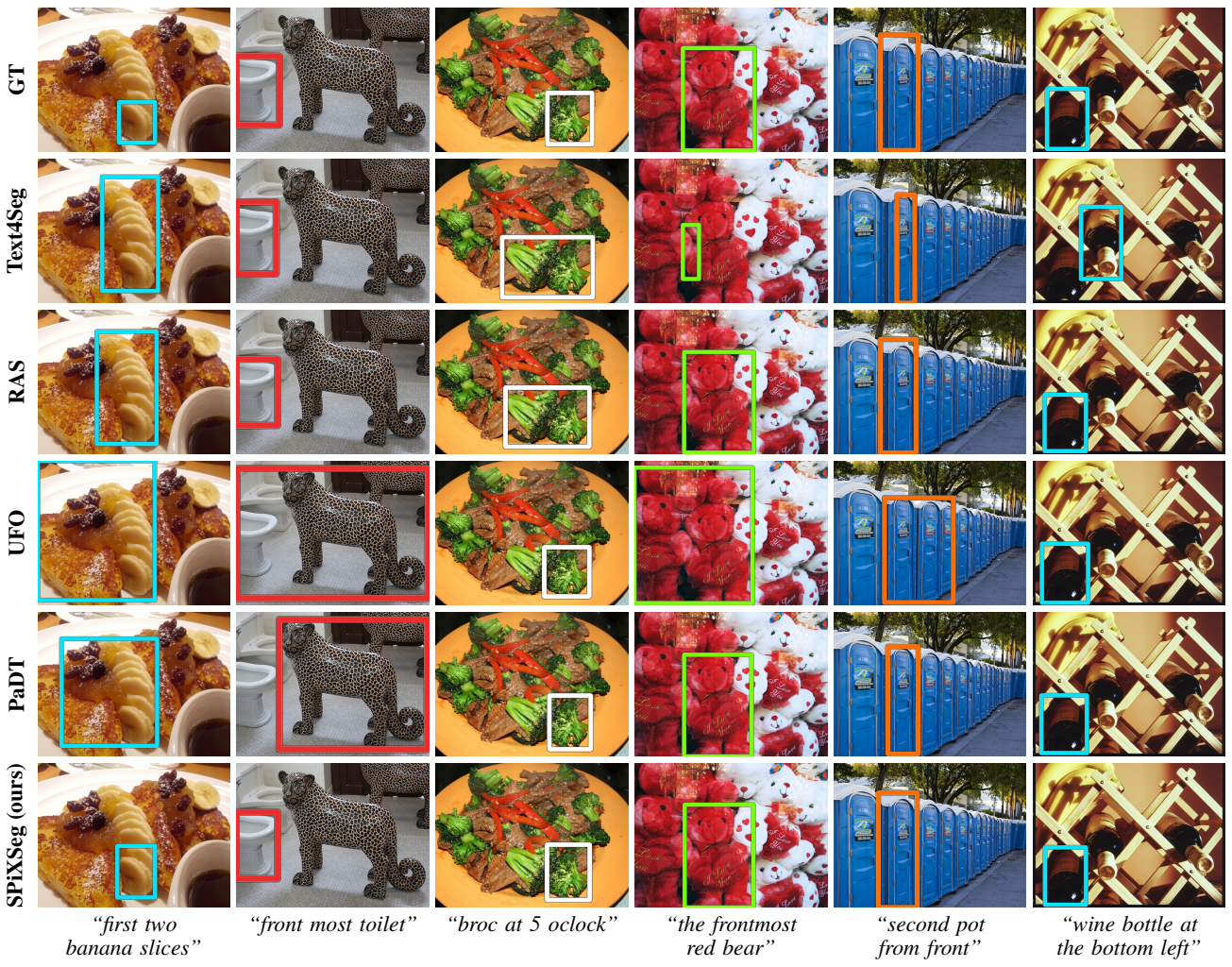}}
    \caption{\textbf{REC.} Boxes are tight around each method's predicted mask (ground truth: around the ground-truth mask), as in Table~\ref{tab:rec}; colours vary per image for visibility only.}
    \label{fig:rec_compare}
  \end{subfigure}
  \caption{\textbf{Qualitative comparison} with representative baselines on RefCOCO.}
  \label{fig:res_rec_compare}
\end{figure*}

\section{Results and Experiments}\label{sec:results}
In this section, we present the performance of our proposed method across tasks such as referring segmentation (RES), referring comprehension (REC), and semantic segmentation. We follow same benchmarking protocols as Text4Seg~\citep{lantext4seg} for each task, and compare against both specialized segmentation models and generalist segmentation models with comparable parameter counts. Our method demonstrates competitive performance across all benchmarks, validating the effectiveness of our approach. We use LLaVA-1.5-7B as the main MLLM with Vicuna-7B as the transformer backbone and CLIP vision tower for fair comparisons; however we also show results with with InternVL2.5-8B (InternLM2.5-7B~\cite{cai2024internlm2}). Further, we use off-the-shelf SAM3.1~\cite{carion2025sam3segmentconcepts} to obtain superpixels. More details regarding superpixel extraction, training and additional experiments are provided in Supplementary material.

\paragraph{Baselines.} We compare our method against a range of baselines, categorised into two groups: (i) specialised segmentation models, which are designed explicitly for segmentation tasks and often rely on task-specific architectures or training recipes; and (ii) generalist segmentation models, which are built on top of LLMs and aim to perform segmentation through in-context learning, dedicated decoder heads, or other architectural adaptations. For semantic segmentation, we include CLIP-based specialists~\cite{liang2023ovseg,xu2023san,sed,fcclip,maftplus,catseg} and the MLLM-based LaSagnA~\cite{wei2024lasagna}, HiMTok~\cite{wang2025himtok} and Text4Seg~\cite{lantext4seg}.

For fair comparison, we mainly include generalist models with 8B parameters or fewer. The exception is RAS, which we also report with its 13B model, since its 7B model reports only the average cIoU.

\paragraph{Datasets.} For semantic segmentation, we evaluate on three datasets: ADE20K~\citep{zhou2017ade20k,zhou2019ade20k} (150 classes), Pascal Context~\citep{mottaghi_cvpr14_pascal_context} (59 classes) and Pascal VOC~\citep{everingham2010pascalvoc} (20 classes). For RES, we evaluate on the three standard benchmarks: RefCOCO, RefCOCO+, and RefCOCOg~\citep{yu2016refcoco}, which contain 80 object categories and varying numbers of referring expressions per image.

\providecommand{\tbd}{\textcolor{red}{\textbf{?}}}%
\providecommand{\chk}[1]{\textcolor{blue}{#1}}%
\providecommand{\todonum}[1]{\textcolor{orange}{#1}}%
\providecommand{\ours}{SPiXSeg}%
\providecolor{vDec}{HTML}{F0F9F9}
\providecolor{vTok}{HTML}{F6F3FC}
\providecolor{vGrid}{HTML}{FDFAE9}
\providecolor{vProp}{HTML}{FEF5EE}
\providecolor{vSpix}{HTML}{E9F6E7}

\begin{table}[t]
\centering
\footnotesize
\setlength{\tabcolsep}{2.0pt}
\resizebox{0.85\linewidth}{!}{%
\begin{tabular}{llcccc}
\toprule
\textbf{Method} & \textbf{LLM / CLIP} & \textbf{A-150} & \textbf{PC-59} & \textbf{PAS-20} & \textbf{Avg.} \\
\midrule
\multicolumn{6}{l}{\textit{CLIP-based specialist models (no LLM)}} \\
OVSeg~\cite{liang2023ovseg} & ViT-B/16 & 24.8 & 53.3 & 92.6 & 56.9 \\
OVSeg~\cite{liang2023ovseg} & ViT-L/14 & 29.6 & 55.7 & 94.5 & 59.9 \\
SAN~\cite{xu2023san} & ViT-B/16 & 27.5 & 53.8 & 94.0 & 58.4 \\
SAN~\cite{xu2023san} & ViT-L/14 & 32.1 & 57.7 & 94.6 & 61.5 \\
SED~\cite{sed} & ConvNeXt-B & 31.6 & 57.3 & 94.4 & 61.1 \\
SED~\cite{sed} & ConvNeXt-L & 35.2 & 60.6 & 96.1 & 64.0 \\
FC-CLIP$^\dagger$~\cite{fcclip} & ConvNeXt-L & 34.1 & 58.4 & 95.4 & 62.6 \\
MAFT+~\cite{maftplus} & ConvNeXt-L & 36.1 & 59.4 & 96.5 & 64.0 \\
CAT-Seg~\cite{catseg} & ViT-B/16 & 31.8 & 57.5 & 94.6 & 61.3 \\
CAT-Seg~\cite{catseg} & ViT-L/14 & 37.9 & 63.3 & 97.0 & 66.1 \\
\midrule
\rowcolor{vDec}\multicolumn{6}{l}{\textit{Decoder after the MLLM}} \\
\rowcolor{vDec}LaSagnA~\cite{wei2024lasagna} & Vicuna-7B & 14.3 & 46.1 & 69.8 & 43.4 \\
\midrule
\rowcolor{vTok}\multicolumn{6}{l}{\textit{Learned mask tokens}} \\
\rowcolor{vTok}HiMTok$^\ast$~\cite{wang2025himtok} & InternLM2.5-7B & \underline{25.0} & 43.9 & \underline{82.0} & \underline{50.3} \\
\midrule
\rowcolor{vGrid}\multicolumn{6}{l}{\textit{Grid cells}} \\
\rowcolor{vGrid}Text4Seg~\cite{lantext4seg} & Vicuna-7B$^\ddagger$ & 16.5 & \underline{52.5} & 76.5 & 48.5 \\
\midrule
\rowcolor{vSpix}\multicolumn{6}{l}{\textit{Superpixels}} \\
\rowcolor{vSpix}\textbf{\ours{} (ours)} & Vicuna-7B$^\ddagger$ & \textbf{29.0} & \textbf{59.5} & \textbf{95.4} & \textbf{61.3} \\
\bottomrule
\end{tabular}%
}
\caption{Open-vocabulary semantic segmentation (mIoU). MLLMs are trained on COCO-Stuff and evaluated zero-shot unless marked. Bold / underline: best / second-best MLLM. $^\ast$Trained with ADE20K and PASCAL Context. $^\dagger$Trained on COCO Panoptic. $^\ddagger$LLaVA-1.5. }
\label{tab:semseg}
\end{table}

\subsection{Referring Expression Segmentation}
\label{sec:results:res}

Referring Expression Segmentation (RES) requires the model to segment an object based on the given image and a natural language description. We evaluate RES performance using class-aware Intersection-over-Union (cIoU) across the three standard benchmarks: RefCOCO, RefCOCO+, and RefCOCOg. Our results are presented in Table~\ref{tab:res}.

With InternVL2.5, SPiXSeg reaches $81.9$, and with LLaVA-1.5 $80.5$, both above the best baselines SAMTok ($79.9$) and UFO ($78.3$). Our method gives the best scores on seven out of eight subsets while SAMTok leads on the remaining one. This illustrates the efficacy of our method which is able to perform at-par with multi-stage training and refinement-based methods. Further, we also observe that SPiXSeg performs better than methods with a decoder head such as PaDT and GLaMM, and than patch-based methods such as UFO and STAMP, even when they are fine-tuned separately for RES, e.g., UFO (RES ft). We attribute this to the simplicity of per-spix classification as opposed to per-pixel classification and native alignment within MLLM which is lost due to mask-specific decoder heads. We present qualitative results in Fig.~\ref{fig:res_compare}. It can be seen that methods like Text4Seg and PaDT which make predictions at patch level lead to coarse masks by construction. While object-proposal based methods like RAS give accurate and precise masks primarily because of the detection model, they fail to reason and often choose incorrect mask as the output. However, we observe that SPiXSeg is able to not only give good quality fine-grained masks but is also able to reason over the superpixels given the text query and recover the correct mask.

\subsection{Referring Expression Comprehension}\label{sec:results:rec}
In referring expression comprehension (REC), the model localizes the referred object with a bounding box, evaluated by Acc@0.5 on the same benchmarks as RES in Table~\ref{tab:rec}. For SPiXSeg, the box is obtained from the predicted mask of the RES model, with no additional training. We observe that SPiXSeg gives the highest average accuracy of $92.2$, followed by SAMTok ($90.2$) and PaDT ($90.1$), and the best score on all subsets. SPiXSeg also performs better than the box-as-text MLLMs, including UFO fine-tuned on each dataset ($88.0$), and the methods that select a detector proposal, e.g., ChatRex ($89.1$). This suggests that the superpixel classification that is trained for segmentation also localizes the object well, without a box being written out. 

\subsection{Semantic Segmentation}\label{sec:results:sem}
In semantic segmentation, the model is tasked with assigning a class label to each pixel in the image. We evaluate this by using mean Intersection-over-Union (mIoU) across all classes on three benchmarks of varying difficulty- ADE20K (150 classes), Pascal Context (59 classes), and Pascal VOC (20 classes). These results are presented in Table~\ref{tab:semseg}. All MLLM-based methods except HiMTok are trained on COCO-Stuff only and evaluated zero-shot on the three benchmarks, which makes the comparison a test of how well the output vocabulary transfers to unseen label spaces and image distributions.

We observe that SPiXSeg gives the highest mIoU among MLLM-based methods on all three benchmarks, with an average of $61.3$ against $50.3$ for HiMTok and $48.5$ for Text4Seg. The gain over Text4Seg, which is the closest comparison since both methods are trained on the same data and use no mask decoder, is $+18.9$ on Pascal VOC, $+12.5$ on ADE20K and $+7.0$ on Pascal Context.  Both methods are given the same number of classes per query, so the difference comes from the unit in which the mask is obtained. In the patch-based format, a class label is predicted for every patch of the grid, and patches lying on object boundaries mix pixels of different classes. A patch is assigned to only one of them, which affects the mIoU of every class that is adjacent to another one. Superpixels follow the image boundaries, so the same decision is made over regions that are mono-semantic and likely of a single class. Methods that use a decoder after the MLLM, e.g., LaSagnA, do not benefit from this, as the semantic prediction has to be translated into a mask by a separate module, and we observe that it also performs below the other methods. HiMTok improves over these on A-150 and PAS-20, but it has been trained with ADE20K and Pascal Context, and hence the comparison on these two benchmarks is not zero-shot. Even so, SPiXSeg is ahead of it on all three benchmarks.

Compared to the CLIP-based specialists, SPiXSeg is at-par with the similar-sized variants like CAT-Seg ViT-B/16, while the larger specialists like CAT-Seg ViT-L/14 lead with average of $66.1$. The gap is the largest on ADE20K, which has the largest class labels. We hypothesize that this is due to the training label space and not the output format, since ADE20K classes without a counterpart in COCO-Stuff must be classified by the MLLM's text understanding alone.

\input{figures/fig_sem_compare_all}

We present qualitative results in Fig.~\ref{fig:sem_compare_all}. It can be seen that Text4Seg often merges different classes into one because of the coarse mask generated. SPiXSeg gives masks that follow the object contours, as each superpixel is assigned entirely to one class. 

\subsection{Additional Results}\label{sec:results:additional}
In this section, we study how the choice of superpixels affects SPiXSeg, compare its inference, time with other paradigms, and present qualitative results on out-of-domain images.

\paragraph{Choice of Superpixels.} Since SPiXSeg segments by selecting superpixels, its accuracy is bounded by how well the superpixels follow the object boundaries. We call the best cIoU reachable by selecting superpixels the `\emph{ceiling}'. Replacing SAM~3.1 with EPOC superpixels~\cite{chen2025epoc} lowers the ceiling by about 9 points as shown in Table ~\ref{tab:spix_compare}, yet SPiXSeg reaches a similar share of it (92.4 vs.\ 93.8\%) and selects nearly the same superpixels as the ceiling with either set as shown in Fig.~\ref{fig:spix_compare}.
\providecommand{\tbd}{\textcolor{red}{\textbf{?}}}%
\providecommand{\chk}[1]{\textcolor{blue}{#1}}%
\providecommand{\todonum}[1]{\textcolor{orange}{#1}}%
\providecommand{\ours}{SPiXSeg}%

\begin{table}[t]
\centering
\small
\setlength{\tabcolsep}{5pt}
\resizebox{0.7\columnwidth}{!}{%
\begin{tabular}{llcccc}
\toprule
\textbf{Superpixels} & \textbf{Method} & \textbf{val} & \textbf{testA} & \textbf{testB} & \textbf{Avg.} \\
\midrule
SAM 3.1 & Ceiling & 89.0 & 89.8 & 88.3 & 89.0 \\
 & \ours{} & 84.1 & 86.9 & 79.7 & 83.6 \\
 & \textbf{Share (\%)} & \textbf{94.5} & \textbf{96.8} & \textbf{90.2} & \textbf{93.8} \\
\midrule
EPOC & Ceiling & 81.1 & 83.4 & 75.5 & 80.0 \\
 & \ours{} & 75.0 & 79.7 & 67.3 & 74.0 \\
 & \textbf{Share (\%)} & \textbf{92.4} & \textbf{95.6} & \textbf{89.3} & \textbf{92.4} \\
\bottomrule
\end{tabular}%
}
\caption{\textbf{Superpixel extractors.} cIoU on RefCOCO with up to 64 superpixels per image. Share: SPiXSeg's cIoU as a percentage of the ceiling.}
\label{tab:spix_compare}
\end{table}

\begin{figure}[t]
  \centering
  \begin{minipage}[t]{0.49\linewidth}
    {\scriptsize\bfseries (a) Accuracy vs.\ latency}\\[1pt]
    \includegraphics[width=\linewidth]{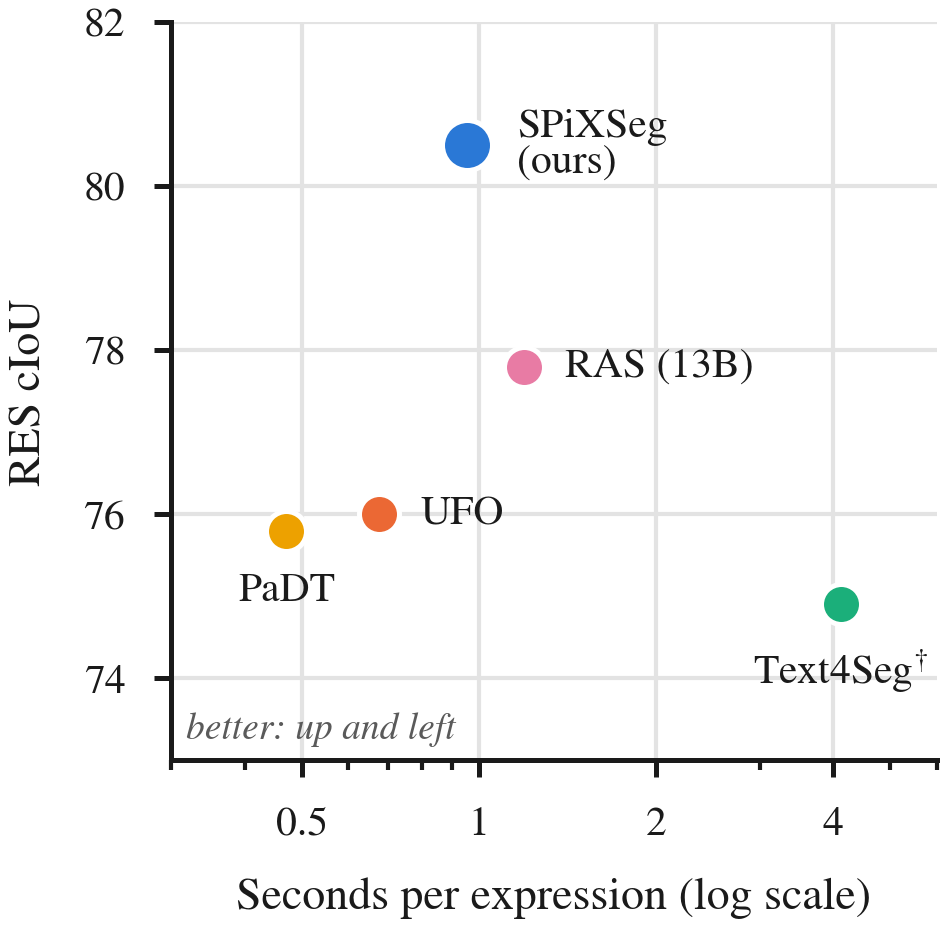}
  \end{minipage}\hfill
  \begin{minipage}[t]{0.495\linewidth}
    {\scriptsize\bfseries (b) Whole-workload time}\\[1pt]
    \includegraphics[width=\linewidth]{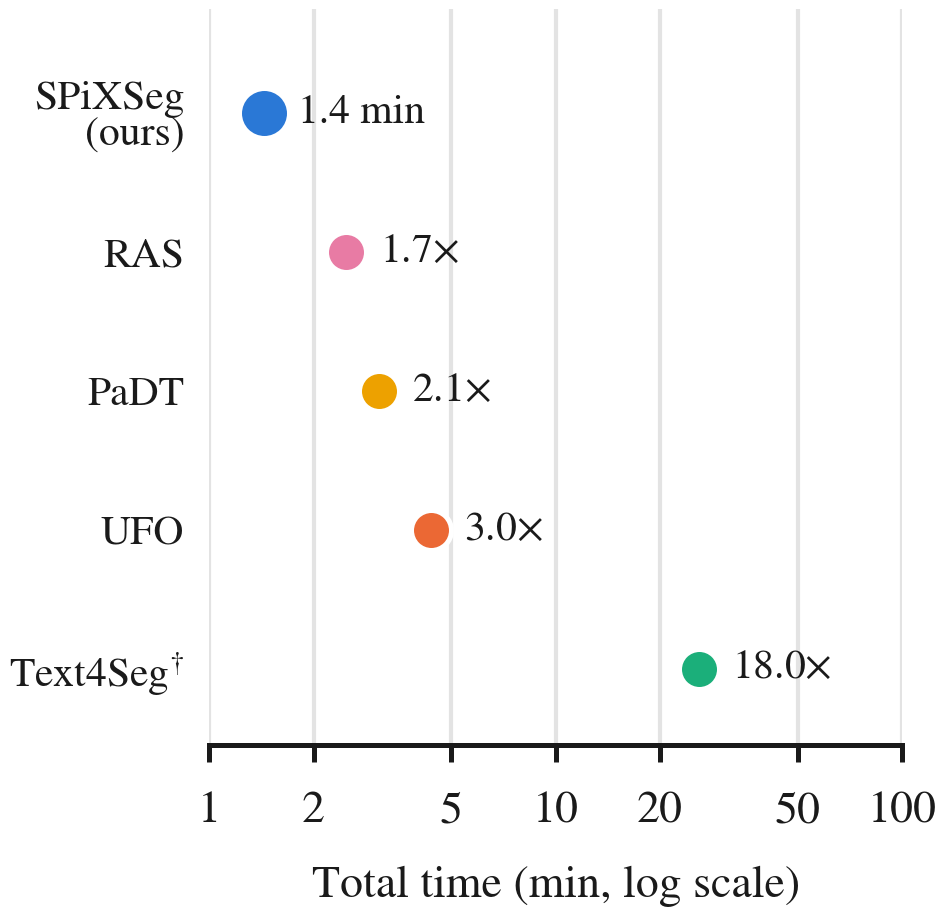}
  \end{minipage}
  \caption{\textbf{Efficiency.} (a) RES cIoU (mean of the 8 splits) against time per expression on a new image (one GPU, batch 1); up and left is better. (b) Total time for all expressions of 50 random RefCOCO val images, with the ratio to SPiXSeg. $^{\dagger}$With SAM-H refinement; without it, 67.2 cIoU at 3.9\,s.}
  \label{fig:tradeoff_workload}
\end{figure}

\input{figures/fig_spix_compare}
\paragraph{Efficiency Comparison.} Fig.~\ref{fig:tradeoff_workload} (a) plots accuracy against the time for one expression on a new image.
SPiXSeg achieves the highest cIoU (80.5) at 0.95\,s, most of which is the one-time superpixel extraction; the model itself takes 0.12\,s. PaDT and UFO are faster on a single new image but 4.5--4.7 points lower in cIoU. In practice, an image is usually queried with several expressions, and SPiXSeg extracts its superpixels only once per image. On a workload of 385 expressions over 50 random RefCOCO val images, as shown in Fig.~\ref{fig:tradeoff_workload} (b), SPiXSeg is the fastest method, finishing in 1.4\,min, $1.7\times$ faster than RAS, $2.1\times$ than PaDT, $3.0\times$ than UFO and $18\times$ than Text4Seg.

\begin{figure}[t]
  \centering
  \resizebox{0.98\columnwidth}{!}{\input{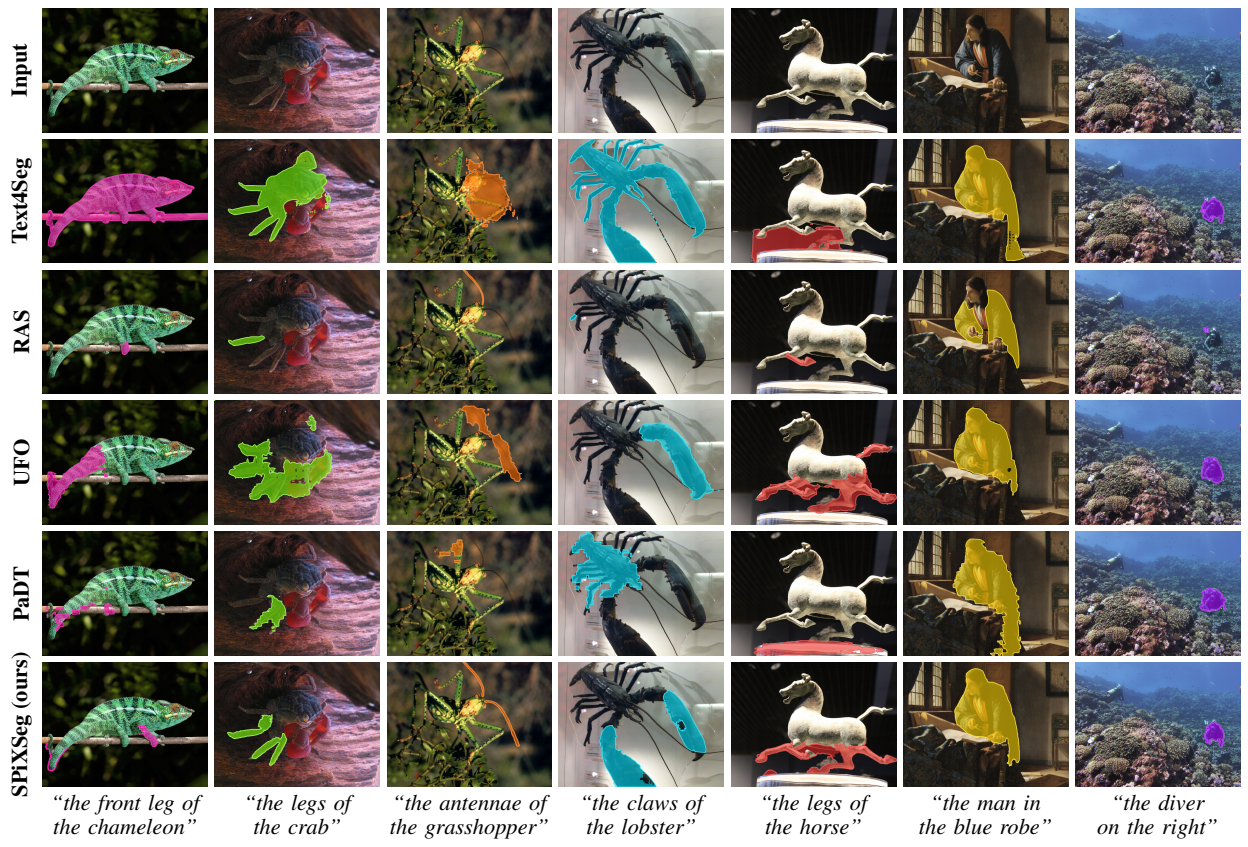}}
  \caption{\textbf{OOD qualitative comparison} of representative baselines and SPiXSeg. Images from Wikimedia Commons; credits in the supplementary material.}
  \label{fig:ood_compare}
\end{figure}

\paragraph{Out-of-domain inputs.} Fig.~\ref{fig:ood_compare} shows images far from COCO, including paintings, sculptures, underwater scenes and close-ups of animals, with expressions that refer to object parts such as legs, claws and antennae. SPiXSeg segments these parts accurately, while the baselines often return the whole object or miss the part.

\section{Conclusion}
In this work, we revisited the vocabulary in which MLLMs express masks for segmentation. Existing vocabularies each have a limitation. Decoder-based methods add per-query overhead, learned mask tokens require multi-stage training, grid-based methods trade boundary precision against the number of decisions, and proposal-based methods only output whole objects. We showed that superpixels satisfy the four properties desired from a vocabulary and proposed SPiXSeg, which recasts MLLM segmentation as superpixel classification. Superpixels are computed once per image and classified by S2M in a single forward pass, without any decoder or refinement. SPiXSeg achieves state-of-the-art results among LLaVA-1.5-7B methods, with a mean cIoU of 80.5 on RefCOCO/+/g, and extends naturally to open-vocabulary semantic segmentation. One limitation is that every mask is a union of superpixels, so accuracy is bounded by the superpixel extraction model and boundaries it misses cannot be recovered. Since SPiXSeg reaches a similar share of this bound with different extractors, we expect better extraction models to act as drop-in replacements that improve masks without architectural changes.

\section*{LLM Usage Statement}

We used LLMs, particularly through the coding agents, in three parts of this work. 
\textbf{Coding:} we used it to write and debug parts of the training, evaluation and analysis code. 
\textbf{Paraphrasing:} we used it to rephrase and edit text that we wrote, for clarity and grammar. 
\textbf{TikZ diagrams:} we used it to write the TikZ code for the method and overview diagrams,
following layouts that we designed. All research ideas, experimental design, results and conclusions are our own.
We reviewed, ran and checked all code; every number in the paper comes from our own experiments; and every qualitative result is an actual output of our model or of a baseline, shown on benchmark images or on openly licensed photographs and artworks. No image or mask was generated or edited with generative AI. We take full responsibility for the content of the paper. The multimodal LLMs we study (LLaVA-1.5 and InternVL2.5) are the subject of the research,
not writing or coding tools, and are described in the main text.

\FloatBarrier 

{    \small
    \bibliographystyle{ieeenat_fullname}
    \bibliography{main}
}

\clearpage
\setcounter{page}{1}
\maketitlesupplementary


\section{Implementation details}
\label{supp:implementation_details}

In this section, we provide the settings and implementation details of the proposed method as described in Sec.~\ref{sec:method} of the main paper. 


\paragraph{Superpixel Extraction.}
As described in the main text, \ours{} method relies on superpixels and treats it as the fundamental unit for segmentation.
However, \ours{} is not tied to a specific superpixel extractor. In our experiments, we use SAM~3.1~\cite{carion2025sam3segmentconcepts} by default and EPOC~\cite{chen2025epoc} as an alternative. With SAM~3.1, we resize the image to $1024{\times}1024$ and prompt the model with a regular (unconditional) $32{\times}32$ grid of points. However, in such regular grid points, neighboring points often produce the same mask, so we remove the near-duplicates with non-maximum suppression at a box IoU of 0.7, and the top 64 masks with the highest predicted IoU are retained.
Since SAM masks are nested, for example, a face lies inside a person; we assign each pixel to exactly one mask going through the masks from the largest to the smallest, so that each part remains a separate superpixel and the larger mask containing it covers only the remaining area. Small uncovered areas and superpixels below 1024 pixels are merged into a neighboring superpixel. This gives at most $N{=}64$ superpixels per image, on RefCOCO we observe the average number of superpixels per image to be 50. For EPOC, the superpixels are determined by the threshold on its predicted boundary map, we choose this threshold to be 0.1.

\paragraph{Model.}
For LLaVA-1.5-7B, we take visual features from the penultimate CLIP layer at a resolution of $336^2$. For InternVL2.5-8B, the input is $896^2$, split into four tiles of $448^2$. In both cases, the vision encoder and the projector are frozen, and the LLM is fine-tuned with LoRA of rank 64. The positional MLP and $\mathbf{e}_{spix}$ in Eq.~\eqref{eq:spix_token} and the scalars $\alpha$ and $\beta$ in Eq.~\eqref{eq:s2m_score} are trained from scratch.

\paragraph{Training.}
We train the pipeline using losses mentioned in Eq.~\eqref{eq:dice} and Eq.~\eqref{eq:focal}. One of the edge cases while training is when none of the superpixels have more than half of its area inside the ground truth, in that case, we use the one with the largest overlap as the positive. 
The losses of Eq.~\eqref{eq:dice} and Eq.~\eqref{eq:focal} have equal weight of 1.0. We train with AdamW at a learning rate of $10^{-4}$
without weight decay, an effective batch of 64, linear warm-up over the first 3\% of steps and cosine decay after
it, gradient clipping at 0.3 and \texttt{bf16}. Since many expressions refer to spatial locations as `left' and `right', the images are resized to a square without flips or crops. For RES, we train on the RefCLEF and RefCOCO/+/g train splits (430K expressions similar to Text4Seg~\cite{lantext4seg}) for 10 epochs with LLaVA-1.5-7B and for 5 epochs with InternVL2.5-8B. During inference, if none of the superpixel reaches $p_i > 0.5$, we keep the one with the highest score.

\paragraph{Other tasks.}
For Generalized RES (gRES), we train from the base MLLM for 5 epochs on gRefCOCO and add a binary cross-entropy term on $\max_i s_i$, which tells whether the expression has a target at all. Unlike in RES, an empty selection is kept and read as `no target'. For ReasonSeg, we start from the RES-finetuned model model and fine-tune it for 2{,}000 steps on batches drawn from data-mix of RES (35\%), VQA on the COCO images of LLaVA-665K (25\%, one $448^2$ tile, text loss), LISA++ (25\%) and the ReasonSeg train split (15\%). 
For semantic segmentation, we train the model for 5 epochs with up to 63 class names per prompt.

\section{Additional experiments}
\label{supp:additional_experiments}
\subsection{Generalized Referring Segmentation}
\label{supp:generalized_referring_segmentation}
Generalized referring segmentation (gRES) refers to task where the expression points to multiple or none of the subjects present in the image.
gRefCOCO~\cite{liu2023gres} contains such expressions that refer to several objects or to none. We train \ours{} on its train split for 5 epochs, with LLaVA-1.5-7B and InternVL2.5-8B. An extra loss on the highest superpixel score teaches the model whether the target exists; if no superpixel is selected, the predicted mask is empty. Table~\ref{tab:supp_gres} presents our results. With InternVL2.5-8B, \ours{} performs the best in every split, and with LLaVA-1.5 our method is second best. We present and compare qualitative examples in  Fig.~\ref{fig:supp_gres_ivl}. It can be seen that our method is able to segment out precise objects and is able to avoid segmentation where the referred object is not present in challenging scenarios.

\providecommand{\tbd}{\textcolor{red}{\textbf{?}}}%
\providecommand{\ours}{SPiXSeg}%
\providecolor{vDec}{HTML}{F0F9F9}
\providecolor{vTok}{HTML}{F6F3FC}
\providecolor{vGrid}{HTML}{FDFAE9}
\providecolor{vProp}{HTML}{FEF5EE}
\providecolor{vSpix}{HTML}{E9F6E7}

\begin{table*}[t]
\centering
\small
\setlength{\tabcolsep}{4.0pt}
\begin{tabular}{llcccccccccc}
\toprule
& & \multicolumn{3}{c}{\textbf{val}} & \multicolumn{3}{c}{\textbf{testA}} & \multicolumn{3}{c}{\textbf{testB}} & \\
\cmidrule(lr){3-5}\cmidrule(lr){6-8}\cmidrule(lr){9-11}
\textbf{Method} & \textbf{LLM} & \textbf{gIoU} & \textbf{cIoU} &\textbf{ N-acc} & \textbf{gIoU} & \textbf{cIoU} & \textbf{N-acc} & \textbf{gIoU} & \textbf{cIoU} & \textbf{N-acc} & \textbf{Avg. cIoU} \\
\midrule
\multicolumn{12}{l}{\textit{Specialist models}} \\
LAVT~\cite{yang2024language} & -- & 58.4 & 57.6 & 49.3 & 65.9 & 65.3 & 49.3 & 55.8 & 55.0 & 48.5 & 59.3 \\
ReLA~\cite{liu2023gres} & -- & 63.6 & 62.4 & 56.4 & 70.0 & 69.3 & 59.0 & 61.0 & 59.9 & 58.4 & 63.9 \\
\midrule
\rowcolor{vDec}\multicolumn{12}{l}{\textit{Decoder after the MLLM}} \\
\rowcolor{vDec}LISA (gRES ft)~\cite{lai2024lisa} & Vicuna-7B & 61.6 & 61.8 & 54.7 & 66.3 & 68.5 & 50.0 & 58.8 & 60.6 & 51.9 & 63.6 \\
\rowcolor{vDec}GSVA (gRES ft)~\cite{xia2024gsva} & Vicuna-7B & 66.5 & 63.3 & 62.4 & 71.1 & 69.9 & 65.3 & 62.2 & 60.5 & 60.6 & 64.6 \\
\midrule
\rowcolor{vTok}\multicolumn{12}{l}{\textit{Learned mask tokens}} \\
\rowcolor{vTok}HiMTok$^\S$~\cite{wang2025himtok} & InternLM2.5-7B$^\star$ & 72.1 & 70.4 & -- & 73.5 & 74.9 & -- & \underline{71.7} & \underline{72.0} & -- & 72.4 \\
\rowcolor{vTok}SAMTok~\cite{zhou2026samtok} & Qwen2.5-3B & 70.5 & 68.2 & 60.5 & 73.8 & 73.0 & 58.8 & 65.9 & 65.4 & 55.3 & 68.9 \\
\midrule
\rowcolor{vGrid}\multicolumn{12}{l}{\textit{Grid cells}} \\
\rowcolor{vGrid}Text4Seg (grid + CRF)~\cite{lantext4seg} & Vicuna-7B$^\ddagger$ & 69.1 & 64.7 & -- & 69.9 & 70.8 & -- & 62.1 & 62.3 & -- & 65.9 \\
\rowcolor{vGrid}Text4Seg (grid + SAM-H)~\cite{lantext4seg} & Vicuna-7B$^\ddagger$ & 73.6 & 67.9 & -- & 74.1 & 72.8 & -- & 66.1 & 64.8 & -- & 68.5 \\
\rowcolor{vGrid}Text4Seg (grid + CRF)~\cite{lantext4seg} & InternLM2.5-7B$^\diamond$ & 70.0 & 66.1 & -- & 69.4 & 70.9 & -- & 63.1 & 64.1 & -- & 67.0 \\
\rowcolor{vGrid}Text4Seg (grid + SAM-H)~\cite{lantext4seg} & InternLM2.5-7B$^\diamond$ & 74.4 & 69.1 & -- & 75.1 & 73.8 & -- & 67.3 & 66.6 & -- & 69.8 \\
\midrule
\rowcolor{vProp}\multicolumn{12}{l}{\textit{Object proposals}} \\
\rowcolor{vProp}RAS~\cite{cao2025RAS} & Vicuna-13B$^\ddagger$ & 68.9 & 64.4 & 57.2 & 74.8 & 74.4 & 54.9 & 66.7 & 64.5 & 54.1 & 67.8 \\
\rowcolor{vProp}RAS (gRES ft)~\cite{cao2025RAS} & Vicuna-13B$^\ddagger$ & 74.6 & 70.5 & 69.1 & 77.5 & 77.0 & 64.6 & 69.4 & 67.9 & 62.9 & 71.8 \\
\midrule
\rowcolor{vSpix}\multicolumn{12}{l}{\textit{Superpixels}} \\
\rowcolor{vSpix}\textbf{\ours{} (ours)} & Vicuna-7B$^\ddagger$ & \underline{77.7} & \underline{71.6} & \underline{70.5} & \underline{78.9} & \underline{78.4} & \underline{66.0} & 71.3 & 70.2 & \underline{63.0} & \underline{73.4} \\
\rowcolor{vSpix}\textbf{\ours{} (ours)} & InternLM2.5-7B$^\star$ & \textbf{81.4} & \textbf{74.5} & \textbf{75.3} & \textbf{81.2} & \textbf{80.2} & \textbf{70.6} & \textbf{73.6} & \textbf{72.3} & \textbf{65.6} & \textbf{75.6} \\
\bottomrule
\end{tabular}
\caption{Generalized referring expression segmentation on gRefCOCO~\cite{liu2023gres}, grouped by vocabulary as in Table~\ref{tab:res}. N-acc: accuracy on expressions with no target (a dash where the paper does not report it). Avg.: mean cIoU over the three splits. gRES ft: fine-tuned on gRefCOCO. Bold: best per column; underline: second best. $^\S$Numbers as reported by SAMTok. $^\ddagger$LLaVA-1.5. $^\star$InternVL2.5-8B. $^\diamond$InternVL2-8B.}
\label{tab:supp_gres}
\end{table*}

\subsection{Reasoning Segmentation}
\label{supp:reasoning_segmentation}
Reasoning Segmentation (ReasonSeg)~\cite{lai2024lisa} refers to the task where the referred object is not directly mentioned and the model has to reason out the relevant object for segmentation.
For ReasonSeg, we fine-tune our RES model on a mix of RES, VQA (the COCO part of LLaVA-665K), LISA++ and the ReasonSeg train split. We present and compare our performance in Table~\ref{tab:supp_reasonseg}. It can be seen that \ours{} performs best on val gIoU and on short queries, and is second best on test gIoU after UFO, which is trained as a generalist with extra reasoning data. We present the qualitative results in Fig.~\ref{fig:supp_reasonseg_ivl}. It can be seen that \ours{} is able to reason the referred object and segment it out precisely.

\providecommand{\ours}{SPiXSeg}%
\providecolor{vDec}{HTML}{F0F9F9}%
\providecolor{vTok}{HTML}{F6F3FC}%
\providecolor{vGrid}{HTML}{FDFAE9}%
\providecolor{vProp}{HTML}{FEF5EE}%
\providecolor{vSpix}{HTML}{E9F6E7}%

\begin{table}[t]
\centering
\small
\setlength{\tabcolsep}{3.0pt}
\resizebox{\columnwidth}{!}{%
\begin{tabular}{llcccccc}
\toprule
& & \multicolumn{2}{c}{\textbf{val}} & \multicolumn{2}{c}{\textbf{test}} & \textbf{short} & \textbf{long} \\
\cmidrule(lr){3-4}\cmidrule(lr){5-6}\cmidrule(lr){7-7}\cmidrule(lr){8-8}
\textbf{Method} & \textbf{LLM} & \textbf{gIoU} & \textbf{cIoU} & \textbf{gIoU} & \textbf{cIoU} & \textbf{gIoU} & \textbf{gIoU} \\
\midrule
\rowcolor{vDec}\multicolumn{8}{l}{\textit{Decoder after the MLLM}} \\
\rowcolor{vDec}LISA (ft)~\cite{lai2024lisa} & Vicuna-7B & 52.9 & 54.0 & 47.3 & 48.4 & 40.6 & 49.4 \\
\rowcolor{vDec}LISA (ft)~\cite{lai2024lisa} & Vicuna-7B$^\ddagger$ & \underline{61.3} & \underline{62.9} & 55.6 & 56.9 & 48.3 & 57.9 \\
\rowcolor{vDec}LaSagnA~\cite{wei2024lasagna} & Vicuna-7B & -- & 47.2 & -- & -- & -- & -- \\
\rowcolor{vDec}UFO~\cite{tang2026ufo} & Vicuna-7B$^\ddagger$ & -- & -- & 58.8 & -- & 46.5 & 62.7 \\
\rowcolor{vDec}UFO~\cite{tang2026ufo} & InternLM2.5-7B$^\star$ & -- & -- & \textbf{67.0} & -- & \underline{56.2} & \textbf{70.4} \\
\midrule
\rowcolor{vTok}\multicolumn{8}{l}{\textit{Learned mask tokens}} \\
\rowcolor{vTok}HiMTok~\cite{wang2025himtok} & InternLM2.5-7B$^\star$ & 60.7 & \textbf{67.0} & 60.8 & \textbf{66.2} & -- & -- \\
\midrule
\rowcolor{vProp}\multicolumn{8}{l}{\textit{Object proposals}} \\
\rowcolor{vProp}Groundhog~\cite{zhang2024groundhog} & LLaMA2-7B & 56.2 & -- & -- & -- & -- & -- \\
\midrule
\rowcolor{vSpix}\multicolumn{8}{l}{\textit{Superpixels}} \\
\rowcolor{vSpix}\textbf{\ours{} (ours)} & InternLM2.5-7B$^\star$ & \textbf{64.4} & 51.2 & \underline{65.5} & \underline{64.7} & \textbf{61.9} & \underline{65.8} \\
\bottomrule
\end{tabular}}
\caption{Reasoning segmentation on ReasonSeg~\cite{lai2024lisa}, grouped by vocabulary. Short / long: test queries split as in LISA. ft: fine-tuned on the ReasonSeg train split. Bold: best per column; underline: second best. $^\ddagger$LLaVA-1.5. $^\star$InternVL2.5-8B.}
\label{tab:supp_reasonseg}
\end{table}

\section{Error Analysis}
\label{supp:error_analysis}
We present an error analysis of our method in Table~\ref{tab:errors_ivl} to illustrate where does our method lose the maximum achievable accuracy.
Table~\ref{tab:errors_ivl} splits the cIoU that \ours{} (with InternVL2.5-8B) loses on RefCOCO/+/g into errors of the model and the limit set by the superpixels (ceiling). The ceiling is the cIoU of the training targets i.e., the highest achievable cIoU. The ceiling cIoU is 88.9 on average, so 11.1 cIoU is lost to the superpixel segmentation inaccuracy irrespective of our method. 
It can be seen that on top of that, our method loses 7.0 points, of which 3.9 comes from picking the wrong object or missing it, and 2.2 from picking the right object with missing or extra pieces. It can be seen that picking wrong objects costs the most on RefCOCO+ testB (10.8 cIoU), where expressions do not use location words. We provide visualization of each of these errors in Fig.~\ref{fig:error_types}.

\providecommand{\ours}{SPiXSeg}%
\providecolor{eObj}{HTML}{FEF5F5}
\providecolor{ePcs}{HTML}{FFF9EE}
\providecolor{ePart}{HTML}{F2F6FB}
\providecolor{vSpix}{HTML}{E9F6E7}
\providecolor{eObjA}{HTML}{C8102E}\providecolor{ePcsA}{HTML}{D98200}\providecolor{ePartA}{HTML}{2171B5}%
\providecolor{eOkA}{HTML}{2E8B3E}%
\providecommand{\edot}[1]{\textcolor{#1}{\raisebox{0.15ex}{\scriptsize$\bullet$}}\,}%

\begin{table*}[t]
\centering
\small
\setlength{\tabcolsep}{5pt}
\resizebox{0.9\textwidth}{!}{%
\begin{tabular}{lccccccccc}
\toprule
& \multicolumn{3}{c}{\textbf{RefCOCO}} & \multicolumn{3}{c}{\textbf{RefCOCO+}} & \multicolumn{2}{c}{\textbf{RefCOCOg}} & \\
\cmidrule(lr){2-4}\cmidrule(lr){5-7}\cmidrule(lr){8-9}
& \textbf{val} & \textbf{testA} & \textbf{testB} & \textbf{val} & \textbf{testA} & \textbf{testB} & \textbf{val} & \textbf{test} & \textbf{Avg.} \\
\midrule
\multicolumn{10}{l}{\textit{(a) Where the remaining cIoU goes (adds up to 100)}} \\[1pt]
\rowcolor{vSpix} \edot{eOkA}\ours{} cIoU & 84.5 & 87.1 & 81.3 & 80.5 & 85.0 & 74.2 & 80.6 & 81.9 & \textbf{81.9} \\
\rowcolor{eObj} \edot{eObjA}+ wrong or missed object & 2.5 & 0.9 & 3.4 & 5.8 & 2.0 & 10.8 & 4.3 & 3.6 & 3.9 \\
\rowcolor{ePcs} \edot{ePcsA}+ right object, wrong pieces & 1.1 & 0.8 & 2.9 & 1.9 & 1.8 & 2.6 & 2.7 & 2.2 & 2.2 \\
\phantom{\edot{eOkA}}+ other (near the ceiling) & 0.9 & 1.0 & 0.8 & 0.9 & 1.0 & 0.8 & 0.6 & 0.7 & 0.8 \\
\rowcolor{ePart} \edot{ePartA}+ partition limit (100 $-$ ceiling) & 11.0 & 10.2 & 11.7 & 10.9 & 10.3 & 11.6 & 11.8 & 11.6 & \textbf{11.1} \\
\midrule
\phantom{\edot{eOkA}}\textit{Piece ceiling} & \textit{89.0} & \textit{89.8} & \textit{88.3} & \textit{89.1} & \textit{89.7} & \textit{88.4} & \textit{88.2} & \textit{88.4} & \textit{88.9} \\
\midrule
\multicolumn{10}{l}{\textit{(b) Error types (\% of expressions)}} \\[1pt]
\rowcolor{eObj} \edot{eObjA}wrong object & 2.9 & 2.0 & 3.5 & 6.1 & 3.7 & 8.2 & 4.4 & 4.1 & 4.4 \\
\rowcolor{eObj} \edot{eObjA}two objects & 2.5 & 1.5 & 3.5 & 3.1 & 1.6 & 4.6 & 2.9 & 2.9 & 2.8 \\
\rowcolor{eObj} \edot{eObjA}mix of objects & 0.8 & 0.6 & 1.8 & 0.9 & 0.7 & 1.7 & 1.0 & 1.0 & 1.0 \\
\rowcolor{eObj} \edot{eObjA}missed & 1.3 & 0.8 & 2.3 & 1.6 & 0.8 & 3.0 & 2.1 & 1.7 & 1.7 \\
\rowcolor{ePcs} \edot{ePcsA}missing pieces & 5.7 & 5.2 & 7.1 & 5.2 & 5.2 & 6.8 & 6.2 & 5.6 & 5.9 \\
\rowcolor{ePcs} \edot{ePcsA}extra pieces & 2.4 & 2.9 & 3.2 & 2.4 & 2.8 & 2.8 & 3.7 & 3.4 & 3.0 \\
\rowcolor{vSpix} \edot{eOkA}near the ceiling (correct) & 84.5 & 87.0 & 78.6 & 80.8 & 85.2 & 73.0 & 79.7 & 81.4 & 81.3 \\
\bottomrule
\end{tabular}
}%
\caption{Where the errors of \ours{} (InternVL2.5-8B) come from; colours match Fig.~\ref{fig:error_types}. (a) The rows add up to 100 on every split. Piece ceiling: cIoU of the training targets, i.e.\ the superpixels more than half inside the ground truth (the top one if none is); the partition limit is what is lost even with this selection. The model errors are measured by replacing each group of predictions with the training targets, in the order listed. (b) Share of expressions per error type: wrong object, two objects, mix and missed make up ``wrong or missed object'' in (a); missing and extra pieces make up ``right object, wrong pieces''.}
\label{tab:errors_ivl}
\end{table*}

\input{figures/fig_supp_gres_ivl}
\input{figures/fig_supp_reasonseg_ivl}
\input{figures/fig_supp_error_types}

\section{Additional Qualitative Results}
\label{supp:add_qual_results}
\subsection{Referring Expression Segmentation}
\label{supp:referring_expression_segmentation}

Table~\ref{tab:res} in the main paper compares \ours{} with prior methods on RefCOCO/+/g. Here, we present more qualitative examples with InternVL2.5-8B backbone in Fig.~\ref{fig:supp_res_ivl}.

\input{figures/fig_supp_res_ivl}

\subsection{Out-Of-Domain Expressions}
\label{supp:out_of_domain_expressions}

Fig.~\ref{fig:ood_compare} of the main paper presented expressions on OOD images that are far from COCO training data. 
We provide more such examples in Fig.~\ref{fig:supp_ood_compare}. These images have no ground truth, and we show expressions on which the baselines struggle.

\input{figures/fig_supp_ood_compare}

\subsection{Open-Vocabulary Semantic Segmentation}
\label{supp:open_vocabulary_semantic_segmentation}

We presented zero-shot qualitative results after training on COCO-Stuff in Table~\ref{tab:semseg}. We provide more qualitative results in Fig.~\ref{fig:supp_sem_ade} for ADE20K-150.

\input{figures/fig_supp_sem_ade}

\subsection{Region Budget}
\label{supp:region_budget}
\ours{} is trained with at most $K{=}64$ superpixels per image, but $K$ can be changed at test time by running the superpixel extraction with a smaller limit, without retraining the model. Fig.~\ref{fig:supp_budget_qual} shows the predictions for $K = 16, 32, 48$ and $64$, together with the ceiling, i.e. the best performance that any selection of the superpixels can reach. With few superpixels, the target is merged with its surroundings, resulting in coarser superpixels, so even the best selection is poor; at $K{=}16$ the ceiling is at most 10 IoU in all four examples. As $K$ grows, the target is covered by its own superpixels and the ceiling rises. From $K{=}32$ on, our prediction reaches the ceiling in every example, so the remaining errors at small $K$ come from the superpixels and not from their classification.

\section{Image credits}
\label{supp:image_credits}

{\footnotesize\centering
\setlength{\tabcolsep}{3pt}
\renewcommand{\arraystretch}{0.95}
\resizebox{\columnwidth}{!}{%
\begin{tabular}{@{}lp{3.7cm}l@{}}
\toprule
Image & Author & Licence \\
\midrule
\multicolumn{3}{@{}l}{\textit{Fig.~\ref{fig:ood_compare}}} \\
\href{https://commons.wikimedia.org/wiki/File:Panther_chameleon_(Furcifer_pardalis)_male_Nosy_Be.jpg}{Chameleon} & \raggedright Charles J. Sharp & CC BY-SA 4.0 \\
\href{https://commons.wikimedia.org/wiki/File:Red-clawed_crab.jpg}{Crab} & \raggedright Björn König & CC BY-SA 3.0 \\
\href{https://commons.wikimedia.org/wiki/File:Grasshopper_insect_macro_photo.jpg}{Grasshopper} & \raggedright Biggins Dean, U.S. Fish and Wildlife Service & Public domain \\
\href{https://commons.wikimedia.org/wiki/File:European_lobster_(Homarus_gammarus)_-_NHM_London.jpg}{Lobster} & \raggedright Federicofuentes & CC BY-SA 3.0 \\
\href{https://commons.wikimedia.org/wiki/File:Eastern_Han_Bronze_Galloping_Horse_(10094846996).jpg}{Horse sculpture} & \raggedright Gary Todd & CC0 \\
\href{https://commons.wikimedia.org/wiki/File:Johannes_Vermeer_-_The_Geographer_-_Google_Art_Project.jpg}{Painting (Vermeer)} & \raggedright Johannes Vermeer & Public domain \\
\href{https://commons.wikimedia.org/wiki/File:NOAA_divers_survey_coral_reef_Rose_Atoll_southwest_2023.png}{Divers} & \raggedright NOAA Fisheries/Hannah Barkley & Public domain \\
\midrule
\multicolumn{3}{@{}l}{\textit{Fig.~\ref{fig:supp_ood_compare}}} \\
\href{https://commons.wikimedia.org/wiki/File:Amphiprion_ocellaris_(Clown_anemonefish)_by_Nick_Hobgood.jpg}{Clownfish} & \raggedright Nick Hobgood & CC BY-SA 3.0 \\
\href{https://commons.wikimedia.org/wiki/File:Ara_ararauna_(Linnaeus_1758).jpg}{Macaw} & \raggedright Michael Gäbler & CC BY 3.0 \\
\href{https://commons.wikimedia.org/wiki/File:Ardea_cinerea_(Grey_heron)_eating_fish.jpg}{Heron} & \raggedright Basile Morin & CC BY-SA 4.0 \\
\href{https://commons.wikimedia.org/wiki/File:Golden-ringed_dragonfly_(Cordulegaster_boltonii)_male.jpg}{Dragonfly} & \raggedright Charles J. Sharp & CC BY-SA 4.0 \\
\href{https://commons.wikimedia.org/wiki/File:Violinist_from_street_in_Montreal.jpg}{Violinist} & \raggedright Wilfredor & CC0 \\
\href{https://commons.wikimedia.org/wiki/File:Pink_whipray_(Pateobatis_fai)_Moorea_2.jpg}{Stingray} & \raggedright Charles J. Sharp & CC BY-SA 4.0 \\
\href{https://commons.wikimedia.org/wiki/File:Catedral,_Teruel,_Espa\%C3\%B1a,_2014-01-10,_DD_62.JPG}{Cathedral} & \raggedright Diego Delso & CC BY-SA 3.0 \\
\href{https://commons.wikimedia.org/wiki/File:Eucomatocera_vittata-Kadavoor-2017-05-23-001.jpg}{Longhorn beetle} & \raggedright Jeevan Jose, Kerala, India & CC BY-SA 4.0 \\
\bottomrule
\end{tabular}}\par}

The out-of-domain images in Fig.~\ref{fig:ood_compare} and Fig.~\ref{fig:supp_ood_compare} are from Wikimedia Commons; we resize them and draw the predicted masks on them. Each image name links to its Commons page.

\input{figures/fig_supp_region_budget}

%


\end{document}